\documentclass[10pt,twocolumn,letterpaper]{article}

\usepackage[pagenumbers]{cvpr} 

\definecolor{cvprblue}{rgb}{0.21,0.49,0.74}
\usepackage[pagebackref,breaklinks,colorlinks,allcolors=cvprblue]{hyperref}
\usepackage{multirow}
\usepackage[dvipsnames,table]{xcolor}
\usepackage{makecell}
\usepackage{graphicx}

\usepackage{epigraph}

\usepackage{titletoc}
\usepackage{cuted} 

\newcommand{\method}{{GroupDiff}\xspace}

\def\paperID{****} 
\def\confName{CVPR}
\def\confYear{2025}

\title{Group Diffusion: Enhancing Image Generation \\ by Unlocking Cross-Sample Collaboration}
\author{
Sicheng Mo$^{1}$ \hspace{0.5mm}
Thao Nguyen$^{2}$ \hspace{0.5mm}
Richard Zhang$^{3}$ \hspace{0.5mm}
Nick Kolkin$^{3}$ \hspace{0.5mm}
Siddharth Srinivasan Iyer$^{3}$ \hspace{0.5mm} \\
Eli Shechtman$^{3}$ \hspace{0.5mm}
Krishna Kumar Singh$^{3}$ \hspace{0.5mm}
Yong Jae Lee$^{3}$ \hspace{0.5mm}
Bolei Zhou$^{1}$ \hspace{0.5mm}
Yuheng Li$^{3}$\\[1mm]
$^{1}$University of California, Los Angeles \hspace{0.5mm}
$^{2}$University of Wisconsin–Madison \hspace{0.5mm}
$^{3}$Adobe Research
\\[1mm]
{\normalsize \url{https://sichengmo.github.io/GroupDiff/}}
}

\begin{document}
\maketitle
\begin{abstract}

In this work, we explore an untapped signal in diffusion model inference. While all previous methods generate images independently at inference, we instead ask if samples can be generated collaboratively. We propose Group Diffusion, unlocking the attention mechanism to be shared across images, rather than limited to just the patches within an image. This enables images to be jointly denoised at inference time, learning both intra and inter-image correspondence.
We observe a clear scaling effect -- larger group sizes yield stronger cross-sample attention and better generation quality. Furthermore, we introduce a qualitative measure to capture this behavior and show that its strength closely correlates with FID.
Built on standard diffusion transformers, our \method achieves up to $32.2\%$ FID improvement on ImageNet-256$\times$256. Our work reveals cross-sample inference as an effective, previously unexplored mechanism for generative modeling.

\end{abstract}    
\section{Introduction}


\begin{figure}[h]
\centering 
\includegraphics[width=1.0\linewidth]{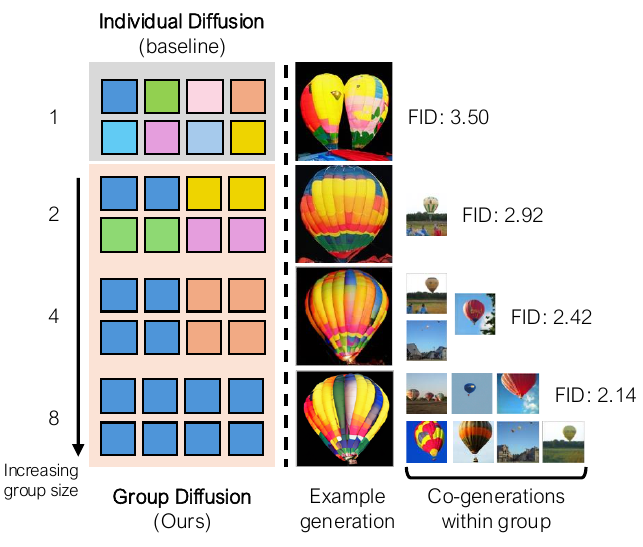}
\vspace{-1em}
\caption{In standard diffusion (top row), samples are generated independently. Our \textbf{\method} uses \textit{cross-sample attention}, enabling samples within a batch to collaborate on a generation. We show selected examples of class-conditional ImageNet generation,
using group sizes $\{1,2,4,8\}$.
We find that the average generation quality improves with larger group size.}
\label{fig:main:teaser}
\end{figure}

\setlength{\epigraphwidth}{.23\textwidth}
\epigraph{``Alone we can do so little; together we can do so much.''}{\textit{Helen Keller}}

During generative model training, network weights are optimized using batches of images to learn an underlying image distribution~\cite{ho2020ddpm,nichol2021iddpm,dhariwal2021adm,rombach2022ldm,stylegan,goodfellow2020gan}. However, at inference time, images are typically generated \textit{independently}. While patches within an image can interact to produce a coherent output, patches across different images are processed separately. This raises an intriguing, unexplored question -- can images and patches \textit{across} a batch collaborate to enhance generation quality collectively?

Following our inquiry, we introduce \textit{Group Diffusion}, which \textit{jointly} denoises a group of samples with the same conditioning. This is enabled using bidirectional attention across samples.
During training, we construct each group by querying semantically or visually similar samples from the training dataset, allowing the attention mechanism to see all patches from within the group.
Then, at test time, we generate images in a batch, allowing images within the batch to aid one another in the diffusion process.

\begin{figure*}[t]
\centering 
\vspace{-0.5em}
\includegraphics[width=0.98\linewidth]{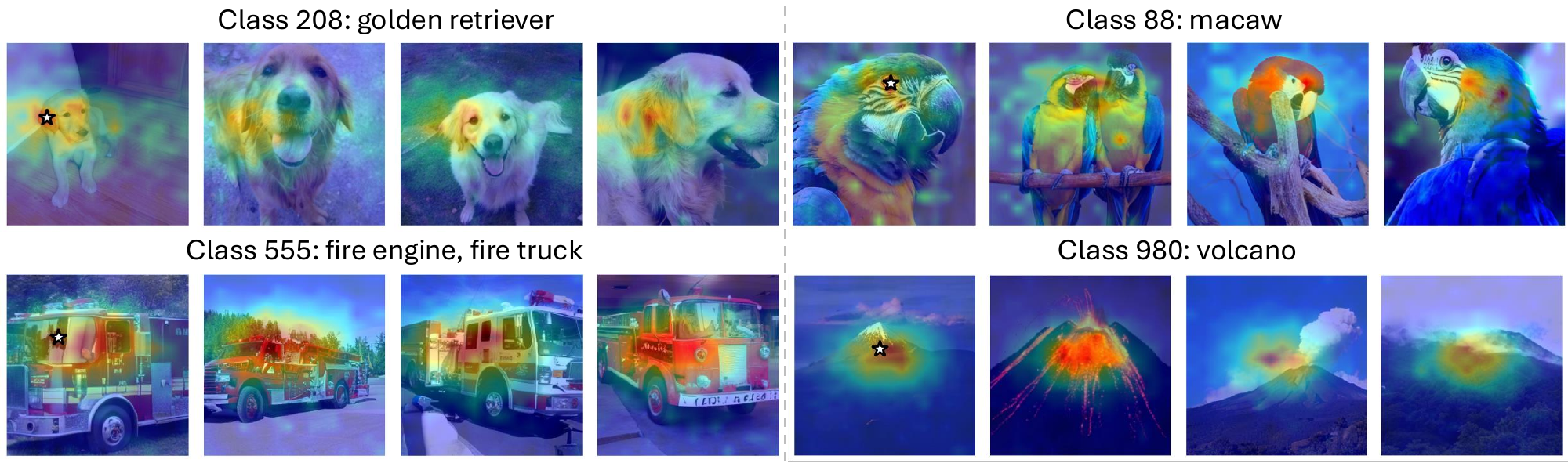}
\vspace{-0.75em}
\caption{\textbf{Attention map visualization}. We show the attention map, using the query point starred on the left, across samples (group size 4), from the second layer. The star refers to the anchor patch. High attention score patches are denoted in red. During the generation process, each image patch is encouraged to attend to similar patches from other images, which enhances the generation quality.  
}
\label{fig:main:attn_map_vis}
\end{figure*}

We observe a clear scaling effect, where increasing the group size strengthens cross-sample attention and consistently improves generation quality, as illustrated in Figure~\ref{fig:main:teaser}. We further analyze the attention patterns across images. As shown in Figure~\ref{fig:main:attn_map_vis}, the group-wise denoising enables each patch to attend to others within the group, allowing the model to learn both intra and inter-image correspondence. Interestingly, we show that generation quality is largely determined by how attention is distributed across samples, with the model assigning higher weights to semantically relevant samples that exert a stronger influence on the final output. We additionally identify a qualitative measure of cross-sample attention whose strength correlates closely with generation quality, providing deeper insight into how group-wise interaction governs the generation process.

We summarize our contribution as follows: 
(1) We present \method{}, a simple yet effective framework that jointly denoise a \textit{group of samples with the same condition} rather than individual images, enabling cross-sample interaction through attention.
(2) A systematic study on \method training and inference behavior, offering insights for better leveraging inter-sample correspondence in image generation. 
(3) Our framework improves generation quality and flexibility over traditional systems;  \eg, integrating \method with SiT yields 20.9\% and 32.2\% better FID when trained from scratch and resumed from a pre-trained checkpoint, respectively.

\section{Related Work}
\smallskip 
\noindent
\textbf{Diffusion models.} Powered by their ability to model complex distributions via iterative denoising, diffusion models have become the leading paradigm for high-fidelity image~\cite{ho2020ddpm,rombach2022ldm,stablediffusion,stablediffusion_3,DALLE2},  video~\cite{openaiSoraCreating,zheng2024opensora,ma2024latte,hong2022cogvideo,yang2024cogvideox,wan2025wan,kong2412hunyuanvideo} and multi-modal concept~\cite{transfusion,shi2024llamafusion,mo2025xfusion} generation.  Besides relying solely on the diffusion objective, recent literature~\cite{repa,leng2025repae,zheng2023maskdit,wu2025reg} explores the alignment between generative modeling and representation learning. REPA~\cite{repa} accelerates diffusion model training by aligning its representation with the pretrained SSL models. REPA-E~\cite{leng2025repae} further leverages the pretrained model's knowledge with additional learnable parameters from the latent encoder. 

Meanwhile, another line of work addresses this potential limitation from the pre-trained vision encoder by aligning cross-layer features to each other (SRA~\cite{jiang2025sra}) or explicitly applying SSL object function on generative model representation (Dispersive Loss~\cite{wang2025disperseloss}). In contrast, \method learns a stronger representation implicitly by allowing group attention to learn both inter and intra-image correspondence.  
This novel approach offers a fresh perspective on integrating diffusion modeling with representation learning.

\smallskip
\noindent
\textbf{Semantic correspondence in diffusion models.} 
Semantic correspondence maps semantically related regions across images, enabling alignment despite changes in appearance or pose.  In addition to its state-of-the-art generation capability, a large-scale pre-trained text-to-image diffusion model~\cite{stablediffusion,stablediffusion_3,LDM} naturally captures such semantic correspondence robustly, which unlocks promising applications in classification~\cite{li2023diff_zero_short_classifier} and segmentation~\cite{tang2022daam,xu2023open_sd_seg,tian2024diffuse_attn_seg} with such features.
Meanwhile, a line of works~\cite{zhang2023a_tale_of_two, luo2023diffusion_hyperfeature} extract high-quality representation from the denoiser by adding different levels of noise and enabling robust cross-image point matching. Follow-up work leverages the global level dense semantic correspondence for image-to-image translation~\cite{tumanyan2023plug_and_play,mo2024freecontrol,lin2024ctrlx, epstein2023diffusion_self_guidance, nguyen2023visual_instruct_inversion} method without additional training. 
Furthermore, there is another line of work that goes beyond single-image generation to multi-view generation~\cite{huang2025mv_adapter}, style-controlled group generation~\cite{sohn2023styledrop}, and video generation~\cite{kara2024rave}, by modeling inter-image correspondence with mutual attention.  
Different from the aforementioned literature, our method explicitly leverages cross-sample relationships to enhance individual sample's quality by jointly denoising all images within a group together, instead of implicitly learning it from individual samples.

\begin{figure*}
    \centering
    \includegraphics[width=1.\linewidth]{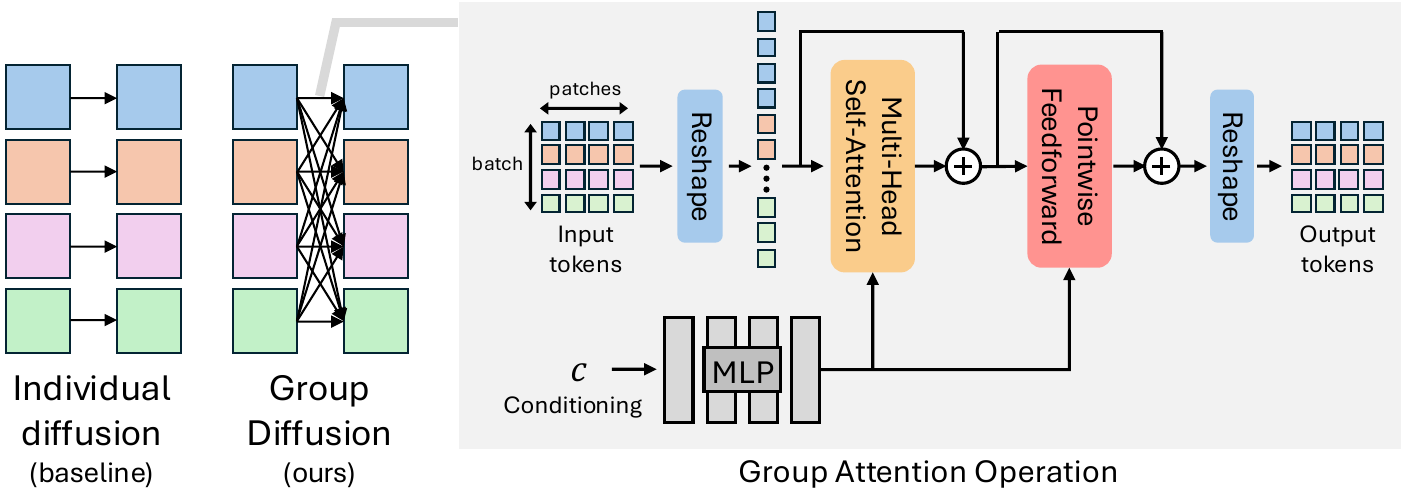}
    \caption{\textbf{Approach.} (Left) Previous approaches generate images independently. We explore Group Diffusion, which allows a set of images to collaborate together during inference time. (Right) Group attention can be implemented simply by reshaping the tokens within a batch, before and after the attention operation.}
    \label{fig:main:appraoch}
\end{figure*}

\smallskip 
\noindent
\textbf{Unified transformer models.} Transformer models~\cite{vaswani2017attention} have unified domain-specific architecture design across language, vision, and audio. It first showcased its strong capability on encoder-decoder and later decoder-only language models in the language domain. ViT~\cite{dosovitskiy2020vit} proposed to convert images to a series of smaller patches to adapt the transformer model to the vision field and find its remarkable scaling capabilities under increasing data, training compute, and data. In the image generative model field, Diffusion Transformer~\cite{peebles2023dit} firstly verified the outstanding scalability of such an architecture, and a similar model design has been further extended to video diffusion models in ~\cite{wan2025wan,kong2412hunyuanvideo,openai2025sora}. Moreover, multi-modal models~\cite{transfusion,Chameleon,shi2024llamafusion,mo2025xfusion} with unified transformer again verified the generaizability of such architecture.  \method benefits from the flexibility of the unified transformer model design by adding multi-image generation capability to the image generative model.

\section{Group Diffusion}

\subsection{Preliminary}

Diffusion models gradually reverse the process of adding noise to an image, starting from a noise vector $\mathbf{x}_{T}$ and progressively generating less noisy samples $\mathbf{x}_{T-1},\mathbf{x}_{T-2},...,\mathbf{x}_{0}$ with learned denoising function $e_\theta$.

The training objective aims to minimize the difference between the predicted and true noise. Specifically, for each time step $t$, the objective is to solve the following denoising problem on the image data $\mathbf{x}$:
\begin{align}
\mathcal{L}_{\text{DM}} = \mathbb{E}_{\mathbf{x}, \epsilon \sim \mathcal{N}(\mathbf{0}, \mathbf{I}), t} \left[ \| \epsilon - e_\theta(\mathbf{x}_t; t, \mathbf{c}) \|_2^2 \right],
\end{align}
where \( \mathbf{x}_t \) is the noisy image at time step \( t \), uniformly sampled from \( \{1, \dots, T\} \), and \( e_\theta(\mathbf{x}_t, t, \mathbf{c}) \) is the denoising function that predicts the noise added to \( \mathbf{x}_t \) conditioned on the time step \( t \) and context \( \mathbf{c} \) (often a text prompt or class label). 

Classifier-free diffusion guidance~\cite{ho2022cfg} enables controlling the trade-off between sample quality and diversity in diffusion models.  It shifts $p_{\theta}(\mathbf{c}|\mathbf{x}_t)$ to assign a higher likelihood to the condition $\mathbf{c}$ without additional classifier. This is implemented by training the diffusion model for both conditional and unconditional denoising and combining the two score estimates at inference time.
Specifically, at inference time, the modified score estimate $\tilde{e}_{\theta}(\mathbf{x}_t,\mathbf{c})$ is extrapolated in the direction towards the conditional $e_{\theta}(\mathbf{x}_t,\mathbf{c})$ and away from the unconditional $e_{\theta}(\mathbf{x}_t,\emptyset)$.
\begin{align}
\tilde{e}_{\theta}(\mathbf{x}_t; t, \mathbf{c}) = e_{\theta}(\mathbf{x}_t;t, \mathbf{c}) + s \cdot \big(e_{\theta}(\mathbf{x}_t; t, \mathbf{c}) - e_{\theta}(\mathbf{x}_;t,\emptyset) \big)
\end{align}

\subsection{Approach}

At the core of our method is the idea of generating multiple images together, so each sample can enhance its generation by selectively learning from other samples, as illustrated in Figure~\ref{fig:main:attn_map_vis}. In our \method, we construct a group with related image data, thus allowing the diffusion model to learn a better representation that can be aided by other samples.
At test time, we generate multiple images, conditioned on the same conditioning $\mathbf{c}$, a setup that aligns well with modern applications, where users typically expect several outputs under the same condition. We follow best practices, adopting the Diffusion Transformer (DiT~\cite{peebles2023dit}) model architecture, which uses an attention mechanism between patches within an image. We simply modify the attention by concatenating the group of image patches together, so that each patch can take other samples into consideration. To ensure that the diffusion model can recognize different image samples, we add the same learnable sample embedding to all patches from a given image.
We formally define the \method method as follows.

\smallskip 
\noindent \textbf{Query method.} Our hypothesis for \method is that images in the same group are related either semantically or visually, and can be used to aid in the denoising process. Thus, we must construct sets of images that are related during training time. Given the image $\mathbf{x}\in \mathbb{R}^{H\times W \times 3}$ and the entire image dataset $\mathcal{D} $, we define the query function $q(\mathbf{x})$ as the following:

\begin{align}
q(\mathbf{x};\mathcal{D}; \tau_{\mathrm{img}}) 
= \left\{ \mathbf{x}_i \in \mathcal{D} \;\middle|\; 
\operatorname{sim}(\mathbf{x}, \mathbf{x}_i) \ge \tau_{\mathrm{img}} 
\right\},
\end{align}

\noindent where $\text{sim}(\cdot)$ returns the image similarity between two images, and $\tau_\text{img}$ is a similarity threshold. 
In practice, we compute the $\text{sim}(\cdot)$ by cosine similarity between image embeddings from pre-trained models like CLIP~\cite{clip} or DINO~\cite{dinov2}.

\smallskip 
\noindent \textbf{\method training.} At each training step, we first construct a group of related images $\mathbf{X}\in\mathbb{R}^{N\times H \times W \times 3}$, including the original image $\mathbf{x}$, by randomly sampling $N-1$ images
from the images returned by query function $q(\mathbf{x}; \mathcal{D}; \tau)$. We use threshold $\tau_{\mathrm{img}}=0.7$ in our experiments, which retrieves a sufficient number of related samples.
For such image group, we first extract their latent with a pre-trained VAE from Stable Diffusion~\cite{stablediffusion}.  To obtain the noisy latent, we sample the timestep independently for each sample but ensure that the variance of the timestep within each group is under the threshold of timestep variation $\sigma_{tv}$. To compute the \textit{group attention}, we first extract the hidden states $h$ from the input $\mathbf{X}$, and then reshape them from $\mathbb{R}^{N\times L \times C} \rightarrow \mathbb{R}^{1\times (N L)\times C}$, where $L$ is the image patch sequence length and $C$ is the channel. After the $Attention(\cdot)$ operation, we reshape the hidden states back.  

In particular, \method enables generating multiple samples in a group by using $L_\text{Group}$ as the loss function as follows: 
\begin{align}
\mathcal{L}_{\text{Group}} = \mathbb{E}_{\mathbf{X}, \mathbf{E} \sim \mathcal{N}(\mathbf{0}, \mathbf{I}), t} \left[ \sum_{i=1}^{N} \left\| \epsilon_i - e_{\theta}(\mathbf{X}; t, \mathbf{c})_i \right\|_2^2 \right]    ,
\end{align}

\noindent where $c$ is the condition and $t$ is the denoising timestep.

\smallskip 
\noindent \textbf{\method inference.} \method enables generating $N$ dependent images following the condition $c$ together at the inference time, instead of $N$ independent image as in previous systems~\cite{peebles2023dit}.
At each timestep, the denoiser predicts two scores: conditional and unconditional. We introduce two variations of our method, \textbf{\method-\textit{f}} and \textbf{\method-\textit{l}}, by flexibly deciding whether to predict the conditional score with group attention or not.  
For \method-\textit{f}, we obtain both scores from group attention and  apply the CFG guidance to combine those scores as follows: 
\begin{align}
\begin{split}
\tilde{e}_{\theta}(\mathbf{X}_{t}; t,\mathbf{c}) &= e_{\theta}(\mathbf{X}_t;t, \mathbf{c}) \\&   + s \cdot (e_{\theta}(\mathbf{X}_t; t, \mathbf{c}) - e_{\theta}(\mathbf{X}_t;t,\emptyset)).
\end{split}
\end{align}
For \method-\textit{l}, only the unconditional score is predicted from group attention. In this case, we obtain $\tilde{e}_{\theta}$ as follows:
\begin{align}
\begin{split}
\tilde{e}_{\theta}(\mathbf{X}_{t}; t,\mathbf{c}) &=  \{ e_\theta(\mathbf{X}_t^i; t, \mathbf{c}) \}_{i=1}^n  \\
& + s \cdot ( \{ e_\theta(\mathbf{X}_t^i; t, \mathbf{c}) \}_{i=1}^n - e_{\theta}(\mathbf{X}_t;t,\emptyset)),
\end{split}
\end{align}
where $\mathbf{X}^{i}_{t}$ is the $i^{th}$ element in group $\mathbf{X}$.

By convention, only 10\% of the data is used to train the unconditional model for generation with CFG~\cite{ho2022cfg}. Since \method-\textit{l} applies the large group size only
to this unconditional model, the remaining 90\% is trained with a group size of one. Thus, most of the training remains identical to standard diffusion, making \method-\textit{l} computationally lightweight compared to \method{}-\textit{f} and
close to baseline systems~\cite{peebles2023dit,ma2024sit}.
Empirically, we find that \method-\textit{l} strikes a good balance between generation quality and computational cost.   
Throughout the paper, we refer to \method{} as \method-\textit{l} unless otherwise specified.

\begin{table*}[t]
\centering
\resizebox{0.85\linewidth}{!}{
\begin{tabular}{ccccccccc}
\toprule
     & \multicolumn{3}{c}{GroupDiff Settings}          & w/o CFG & \multicolumn{2}{c}{w/ CFG} & Cross-Sample & Linear Prob \\
\cmidrule(r){2-4} \cmidrule(lr){5-5}\cmidrule(lr){6-7}
Iter & Model          & Query Method & Noise Var. & FID~$\downarrow$     & FID~$\downarrow$       & cfg-scale      & Attn. Score~$\uparrow$        & Acc.~$\uparrow$         \\
\midrule
800K & C = 1, UC = 1  & -            & 0               & 14.38   & 3.50      & 1.5            & -          & 49.48       \\
\midrule 
800K & \cellcolor{blue!15} C = 4, UC = 4  & Class            & 0               & 14.27   & 3.08      & 1.6            & -     & 62.15       \\
800K & \cellcolor{blue!15} C = 1, UC = 4  & Class            & 0               & 13.22   & 2.81      & 1.6            & -     & 64.44       \\
\midrule 
800K & \cellcolor{green!15} C = 1, UC = 2  & CLIP-L       & 0               & 13.47   & 2.92      & 1.5            & 0.00\%       & 55.33       \\
800K & \cellcolor{green!15} C = 1, UC = 4  & CLIP-L       & 0               & 13.93   & 2.42      & 2.0            & 19.95\%      & 58.83       \\
800K & \cellcolor{green!15}  C = 1, UC = 8  & CLIP-L       & 0               & 13.08   & 2.14      & 2.2            & 51.13\%      & 67.93       \\
800K & \cellcolor{green!15} C = 1, UC = 16 & CLIP-L       & 0               & 13.84   & 1.86      & 2.5            & 56.47\%      & 72.91       \\
\midrule 
800K & C = 1, UC = 4  & \cellcolor{red!15} Random       & 0               & 13.28   & 3.57      & 1.5            & 23.17\%      & -           \\
800K & C = 1, UC = 4  & \cellcolor{red!15} Class        & 0               & 13.22   & 2.81      & 1.6            & 22.51\%      & 64.44       \\
800K & C = 1, UC = 4  & \cellcolor{red!15} CLIP-B       & 0               & 13.47   & 2.51      & 1.9            & 19.20\%      & 61.14       \\
800K & C = 1, UC = 4  & \cellcolor{red!15} CLIP-L       & 0               & 13.93   & 2.42      & 2.0            & 19.95\%      & 58.83       \\
800K & C = 1, UC = 4  & \cellcolor{red!15} SigLIP       & 0               & 13.83   & 2.45      & 2.0            & 19.98\%      & 63.32       \\
800K & C = 1, UC = 4  & \cellcolor{red!15} DINOv2-B       & 0               & 14.40   & 2.51      & 1.9            & 18.45\%      & 63.32       \\
800K & C = 1, UC = 4  & \cellcolor{red!15} DINOv2-L       & 0               & 13.35   & 2.51      & 1.9            & 22.85\%      & 59.16       \\
800K & C = 1, UC = 4  & \cellcolor{red!15} I-JEPA        & 0               & 13.08   & 2.44      & 1.8            & 18.50\%      & 60.50       \\
\midrule 
800K & C = 1, UC = 4  & CLIP-L       & \cellcolor{yellow!15} 0               & 13.93   & 2.42      & 2.0            & 19.95\%      & 58.83       \\
800K & C = 1, UC = 4  & CLIP-L       & \cellcolor{yellow!15} 20              & 13.50   & 2.42      & 2.0            & 21.37\%      & 60.48       \\
800K & C = 1, UC = 4  & CLIP-L       & \cellcolor{yellow!15} 50              & 12.81   & 2.34      & 2.0            & 26.23\%      & 68.91       \\
800K & C = 1, UC = 4  & CLIP-L       & \cellcolor{yellow!15} 100             & 12.78   & 2.32      & 1.9            & 23.33\%      & 62.82            \\
800K & C = 1, UC = 4  & CLIP-L       & \cellcolor{yellow!15} 150             & 13.70   & 2.25      & 2.0            & 24.31\%      & 63.74            \\
800K & C = 1, UC = 4  & CLIP-L       & \cellcolor{yellow!15} 200             & 13.26   & 2.32      & 1.8            & 24.46\%      & 60.03       \\
\bottomrule
\end{tabular}
}

\vspace{-0.5em}
\caption{\textbf{Component-wise analysis} on ImageNet $256 \times 256$ with DiT-XL/2~\cite{peebles2023dit} trained for 800K iterations. All metrics except accuracy (Acc.) are measured with the iDDPM~\cite{nichol2021iddpm} 
sampler with NFE$=250$. For generation results with Classifier-Free Guidance, we search for the optimal guidance scale using an interval of 0.1 and report the one with the optimal FID score. $\uparrow$ and $\downarrow$ indicate whether higher or lower values are better, respectively. C and UC referring to the conditional model and unconditional model,respectively. 
}
\label{tab_ablation}
\end{table*}

\section{Experiments}
We now analyze our proposed \method, beginning with the introduction of the experiment setup and a series of ablation studies on the group settings, followed by observations of the intriguing property and behavior of \method{}. Lastly, we benchmark with previous leading systems. 

\subsection{Setup}

\smallskip 
\noindent
\textbf{Implementation Details.} We strictly follow the DiT~\cite{peebles2023dit} and SiT~\cite{ma2024sit} model architecture/configuration and data process. We train the \method with AdamW optimizer, a constant learning rate of $1\times10^{-4}$, and weight decay $0.01$ on A100 GPUs.
Sampling is performed using the SDE Euler-Maruyama sampler and the iDDPM~\cite{nichol2021iddpm} sampler with $\text{NFE}=250$ when SiT~\cite{ma2024sit} and DiT~\cite{peebles2023dit} are selected as the baseline model, respectively. 
We consistently use a global batch size of 256 when adjusting the group size to ensure a fair comparison across variations and baseline methods.  
Additional implementation details and baseline introduction are provided in the Supplementary. 

\smallskip 
\noindent
\textbf{Datasets and metrics.} Following DiT~\cite{peebles2023dit}, we conduct experiments on ImageNet~\cite{imagenet} and use a pretrained Stable Diffusion VAE with a compression ratio of $8$ to encode each $256\times256$ image into a compressed vector $x \in \mathbb{R}^{32\times32\times4}$. And we report the FID~\cite{heusel2017fid}, Inception Score~\cite{salimans2016is_score}, Precision and Recall~\cite{kynkaanniemi2019improved_precision_and_recall} for measuring the generation quality.

\subsection{Main Properties}

As shown in Table~\ref{tab_ablation}, we discover that \method consistently provides a substantially improved generation performance across various design choices, achieving a much better FID score than the vanilla
model. Below, we provide a detailed analysis of the impact of each component.

\smallskip
\noindent
\colorbox{blue!15}{\textbf{Group model.}} 
The leading diffusion systems usually benefit from Classifier-Free Guidance~\cite{ho2022cfg}, which takes the joint effect with the conditional model and unconditional model. In practice, those two models usually share most model weights besides the condition embedding. We begin the ablation by analyzing the model behavior when applying the Group Attention operation on one or both models.

In this analysis, we use the ImageNet~\cite{imagenet} class label as the query method  to build each group.  We observe that \method consistently outperforms the individual diffusion baseline.  Notably, when only running the \textbf{UC} model in the \method mode, our system further achieves higher generation quality when both the CFG is disabled or enabled, reflected by lower FID. Under this setting, we observe that the condition model's generation capability has also improved when we train only the unconditional model with group attention. We hypothesize that the stronger representation in the UC model implicitly enhances the C model via weight sharing. In later experiments, we set \textit{C=1, UC=N} as the default choice to balance training and inference.

\smallskip
\noindent
\colorbox{green!15}{\textbf{Group size.}} 
We also study the impact of group size in \method. Larger groups generally yield better generation results, as reflected by consistent improvements in FID and feature quality. 
We hypothesize that larger groups offer greater flexibility for finding better patch-level matches, thereby enhancing generation and internal representations. Detailed pattern analysis is provided in Sec.~\ref{sec:pattern}.
In the following experiments, we choose 4 as the group size for fair comparison with baseline methods. 

\begin{figure}[t]
\centering 
\includegraphics[width=0.9\linewidth]{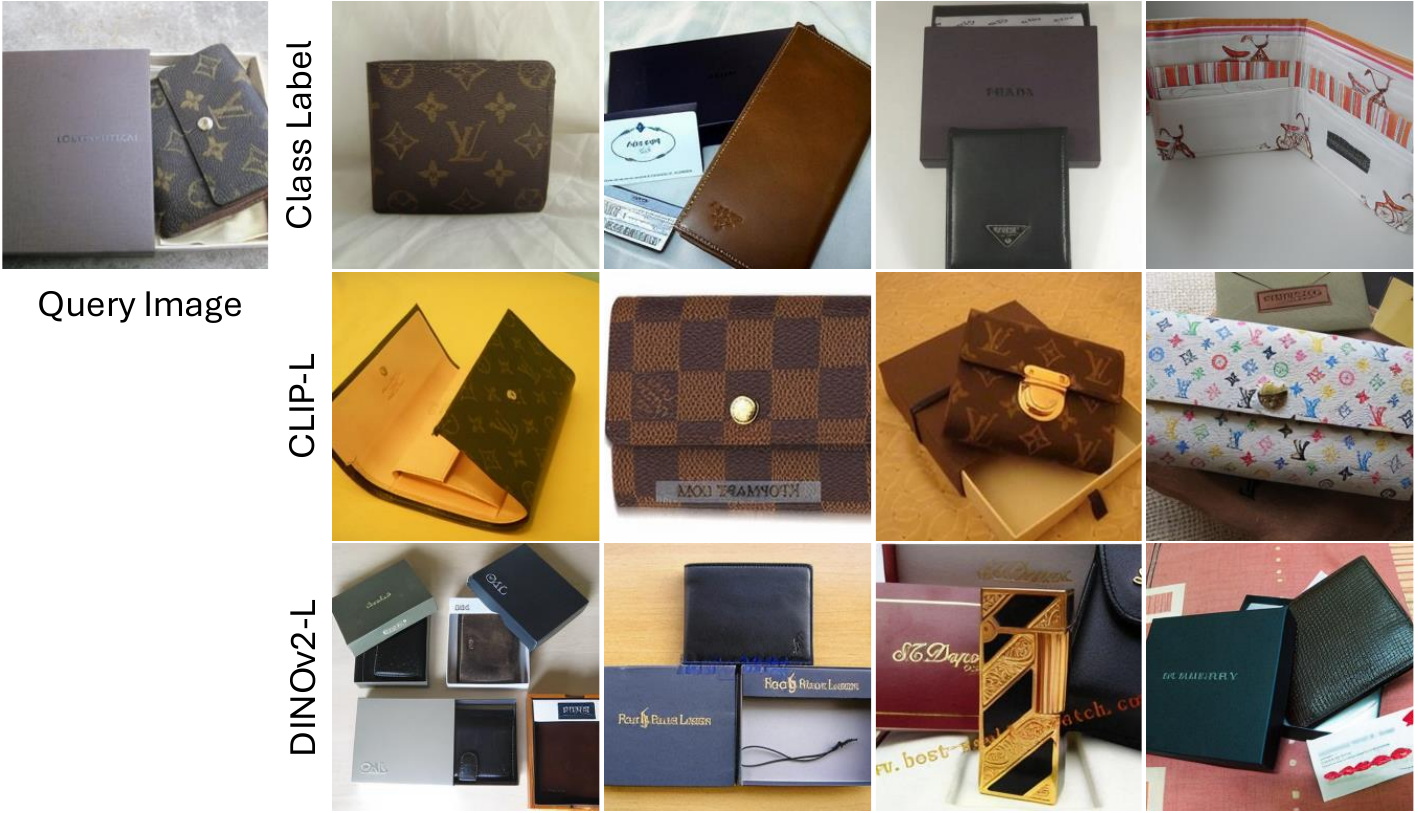}
\vspace{-0.25em}
\caption{\textbf{Comparison of group candidates from different query methods.}
The difference in pretraining settings lead each query method to form distinct groups. We show nearest samples from the ImageNet~\cite{imagenet} training split, with the class label row showing random same-class samples.
}
\label{fig:main:query_results}
\vspace{-1em}
\end{figure}

\begin{figure*}[t]
\centering 
\vspace{-0.5em}
\includegraphics[width=0.94\linewidth]{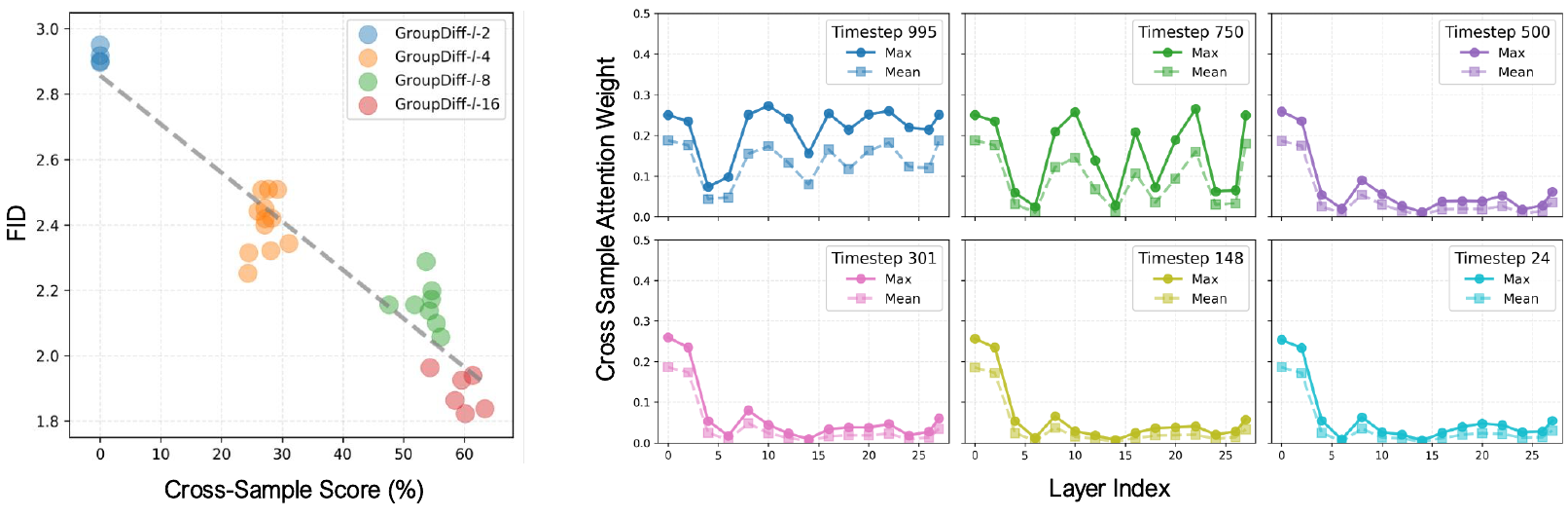}
\vspace{-1.0em}
\caption{
\textbf{Cross-Sample Attention in \method.}
(a) \textbf{FID vs Cross-Sample Score (left).} Our \method shows a strong correlation (0.95) between cross-attention to other samples and generation quality. 
(b) \textbf{Cross-Sample Attention Visualization (right).}
}
\vspace{-1.0em}
\label{fig:main:attn-stats}
\end{figure*}

\smallskip
\noindent
\colorbox{red!15}{\textbf{Group construction method.}} We then investigate the impact of different group construction methods, including random sampling, class-based grouping, and similarity-based retrieval via pre-trained vision encoders. Quantitatively, similarity-based grouping yields the best generation quality, followed by class-based grouping, while random sampling performs the worst (on par with the baseline). This indicates that group attention does not degrade the baseline diffusion model’s performance, even without any bells and whistles. Meanwhile, we hypothesize that image similarity within a group is crucial for strengthening cross-sample interaction. Random groups often contain unrelated samples and thus lack meaningful mutual information, whereas similarity-based retrieval retrieves semantically coherent images, reducing the FID (with CFG) from $3.57$ to around $2.4$. 

Interestingly, Figure~\ref{fig:main:query_results} shows that different pre-trained encoders form visually distinct groups.
For instance, CLIP-L~\cite{clip} tends to cluster semantically similar samples, while DINOv2-B~\cite{dinov2} captures alternative aspects of visual similarity. Nevertheless, their resulting generation quality remains comparable, suggesting that the benefit primarily arises from semantic consistency rather than the specific encoder style. Overall, \method demonstrates strong flexibility and generalization, showing that the quality of the pre-trained encoders does not limit its performance.

\smallskip
\noindent
\colorbox{yellow!15}{\textbf{Group noise-level variation.}} Lastly, we explore the effect of introducing noise-level variance within each group.  Instead of applying the same noise level to the entire group, we restrict the noise levels of the other samples to differ from that of the first sample by up to a specified range, \eg 50 or 200. Prior works~\cite{chen2020simclr,yang2025ldetok} verified that adding different level of noise could be an effective augmentation method for improving representations learning and generation quality.  In our setting, we hypothesize that noisier samples benefit from cleaner ones within the same group, further encouraging cross-sample attention. We find that setting the noise-level variation in the range of $50$ to $200$ yields the best performance, improving both FID and linear probe accuracy while strengthening cross-sample attention.

\subsection{\method Generation Pattern Analysis}
\label{sec:pattern}

After validating the effectiveness of different group settings, we now analyze why and how \method improves generation quality and investigate its unique generation patterns.

\smallskip
\noindent
\textbf{Cross-Sample Attention}. 
To understand why \method improves generation, we examine how cross-sample interaction influences the diffusion process.
At the core of \method{} is cross-sample attention, enabling each patch to establish intra-image and inter-image correspondence across the group.
Figure~\ref{fig:main:attn_map_vis} shows that a patch corresponding to a “dog’s ear” attends to both the same region of its own instance and to similar “ear” regions in other dog images.

To quantitatively measure the cross-sample attention,   we define the image-level \emph{self-attention} as attention assigned to its own patches, and \emph{cross-attention} as attention assigned to patches from other images in the group.
Formally, let image $x_i$ contain patch indices $\mathcal{I}_i$. 
For a query patch $q \in \mathcal{I}_i$ and any key patch $k$, let the attention weight be $\alpha_{qk}$.
We define the image-level cross-attention weight for image $x_i$ as
\begin{align}
\begin{split}
P^{x_i}_{\text{cross}}
= \bigl\{ p^{x_i \rightarrow x_j} \,\big|\, j \neq i \bigr\}, \\
\text{where }
p^{x_i \rightarrow x_j}
= \sum_{q \in \mathcal{I}_i} \sum_{k \in \mathcal{I}_j} \alpha_{qk}.
\end{split}
\end{align}

\begin{figure}[t]
\centering 
\vspace{-0.5em}
\includegraphics[width=0.8\linewidth]{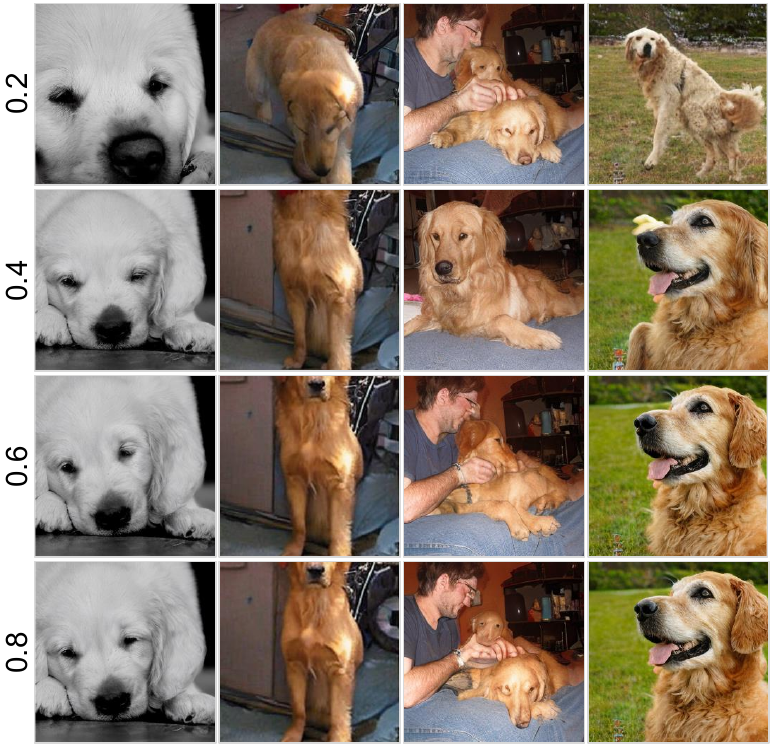}
\vspace{-0.5em}
\caption{\textbf{Controlling \method denoising steps.} We show generated sample examples when \method is turned off after different denoising stages. Stable quality after denoising with \method at early steps.
}
\label{fig:main:qual_over_steps}
\vspace{-1.0em}
\end{figure}

Furthermore, we introduce the \emph{mean cross-attention score} and the 
\emph{max cross-attention score} of image $x_i$ by taking the mean and 
maximum over $P^{x_i}_{\text{cross}}$:
\begin{align*}
 P_{\text{cross-mean}}^{x_i}
= \operatorname{mean}\!\left(P^{x_i}_{\text{cross}}\right),
\qquad
P_{\text{cross-max}}^{x_i}
= \max\!\left(P^{x_i}_{\text{cross}}\right).   
\end{align*}

\textit{Attention over denoising steps.}
To further quantify this effect, we measure cross-sample attention across different denoising steps using the image-level cross-attention score, $P^{x_i}_{\text{cross}}$. For each image, we compute its mean and maximum cross-attention scores, $P_{\text{cross-mean}}^{x_i}$ and $P_{\text{cross-max}}^{x_i}$, and average these statistics over all images in the group. As shown in Figure~\ref{fig:main:attn-stats} (right), both the mean and maximum cross-attention scores gradually decrease as the noise level reduces, indicating that inter-sample information exchange is most active at the early stages of denoising when global structure and semantics are being formed.

To validate this observation, we conduct an intervention experiment by turning off \method after a certain number of denoising steps and continuing the process using the baseline DiT model. As illustrated in Figure~\ref{fig:main:qual_over_steps}, disabling \method{} in the middle or late stages yields little quality degradation, confirming our aforementioned hypothesis. Table~\ref{tab:timestep-ablation} shows that \method could be faster without degraded quality by only applying group attention in the early and middle stages.

\textit{Attention over denoiser layers.} 
We also examine the layer-wise distribution of cross-sample attention. \method{} shows stronger cross-sample attention in the early and final layers, suggesting
that it uses other samples to form global context and later refine details. Table~\ref{tab:layer-ablation} shows that early layers are essential, while late layers have much less impact on \method.
These results indicate \method{} strengthens cross-sample interaction in the early timesteps and shallow layers, leading to improved generation quality.

\begin{table}[h]
\resizebox{0.85\linewidth}{!}{%
\begin{minipage}{0.4\linewidth}
\begin{tabular}{p{0.5\linewidth} c}
\toprule
Methods & FID-10K \\
\midrule
Baseline   & 4.21 \\
~w/t 0.0-0.2 & 4.04 \\
~w/t 0.0-0.4  & 3.92 \\
~w/t 0.0-0.6 & 4.63 \\
\bottomrule
\end{tabular}
\vspace{-0.5em}
\caption{\textbf{Ablation} on group attention timestep.}
\label{tab:timestep-ablation}
\end{minipage}
\hspace{2em}
\begin{minipage}{0.4\linewidth}
\centering
\begin{tabular}{p{0.7\linewidth} c}
\toprule
Methods & FID-10K \\
\midrule
Baseline   & 4.21 \\
~w/o layer 1-9 & 294.38 \\
~w/o layer 10-19 & 5.49 \\
~w/o layer 20-27 & 4.49 \\
\bottomrule
\end{tabular}
\vspace{-0.5em}
\caption{\textbf{Ablation} on group attention layers.}
\label{tab:layer-ablation}
\end{minipage}
} 
\vspace{-1em}
\end{table}

\smallskip
\noindent
\textbf{Cross-sample attention score.}
Under a setting that encourages cross-image attention, we hypothesize two possible operating modes: (i) an evenly distributed mode, where an image spreads attention across all others, and (ii) a neighbor-focused mode, where it primarily attends to its most similar counterpart.
We focus on the latter behavior and quantify its strength using an image-level cross-sample attention score defined as
\begin{align}
S_{\text{cross}} = 
\frac{P_{\text{cross-max}} - P_{\text{cross-mean}}}{P_{\text{cross-max}}},
\end{align}
where \(P_{\text{cross-max}}\) and \(P_{\text{cross-mean}}\) denote the maximum
and mean cross-sample attention from one image to the others in the group.  Intuitively, this score measures how strongly the attention distribution concentrates on the most similar image, normalized by the overall attention magnitude. A score close to 0 indicates a uniform, distributed attention pattern, while a score close to 1 reflects a highly peaked, neighbor-focused attention on a single image.

By varying the query method, noise range, and group size across \method{} variants, we compare their cross-sample attention scores with their FID.  We observe a strong correlation ($r = 0.95$; Fig.~\ref{fig:main:attn-stats} left), showing that more neighbor-focused cross-sample attention leads to higher generation quality. 
Upon closer inspection, several distinct clusters emerge in the plot, primarily corresponding to different group sizes. We find that increasing the group size effectively encourages stronger cross-sample attention behavior, further improving generation quality. 
Moreover, even within each cluster, higher cross-sample attention scores still correlate with lower FID, showing that this interaction reliably reflects generation quality.

\begin{figure}[t]
\centering 
\vspace{-0.5em}
\includegraphics[width=0.85\linewidth]{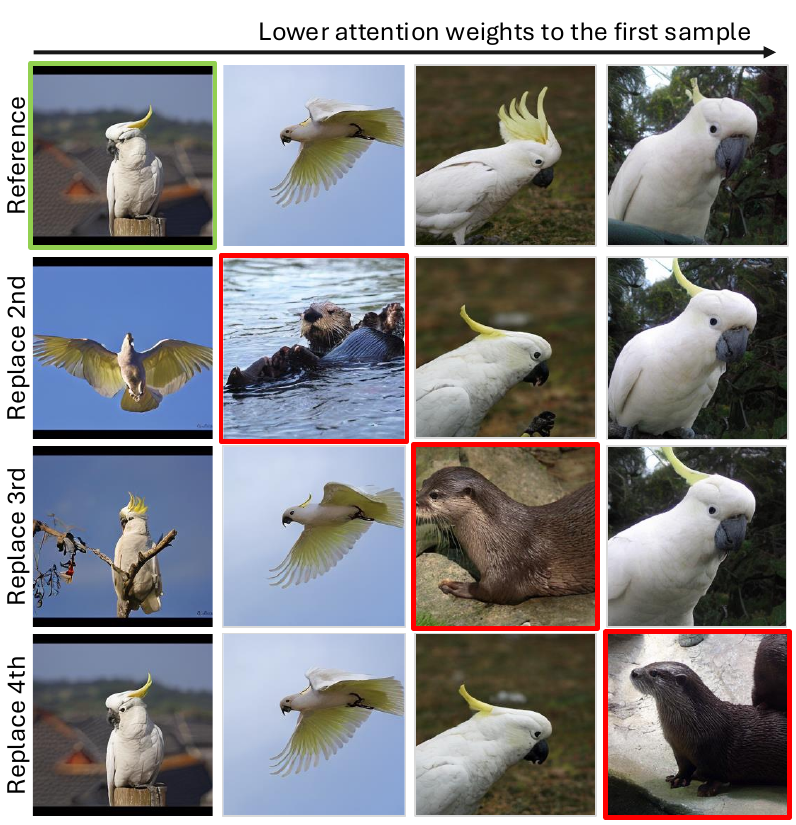}
\vspace{-1em}
\caption{\textbf{Controlling conditions.} The reference group uses class 89, and in each row, one sample’s (red) condition is changed to class 360.}
\label{fig:main:control_class}
\vspace{-1.5em}
\end{figure}

\begin{figure*}[!h]
\centering 
\vspace{-0.5em}
\includegraphics[width=0.9\linewidth]{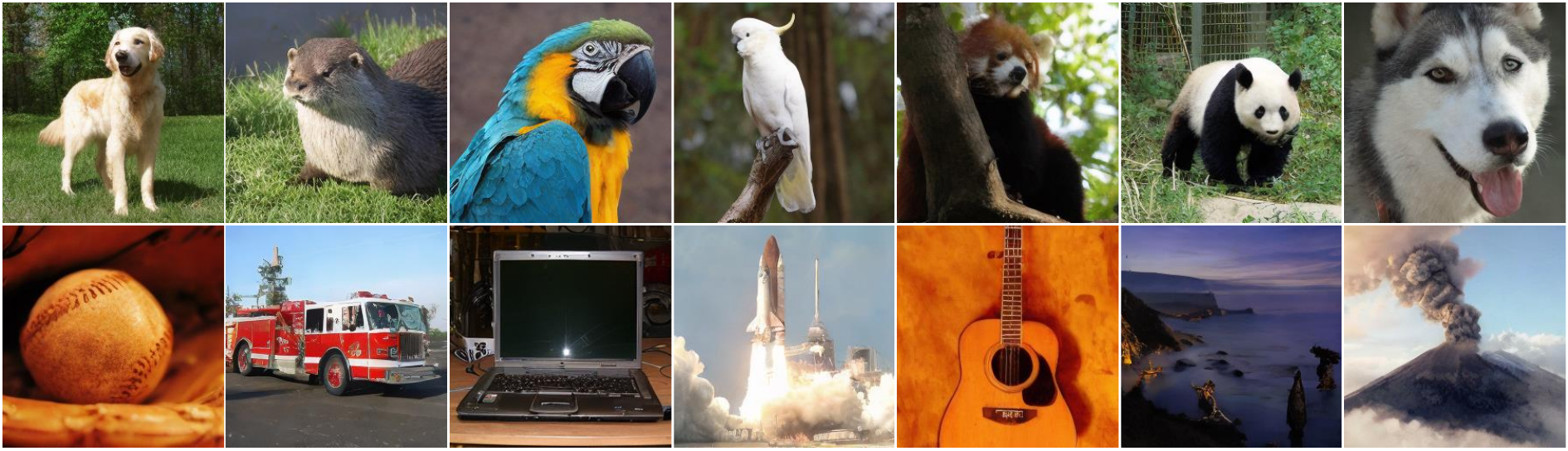}
\vspace{-0.75em}
\caption{\textbf{Qualitative Results.} Examples of class-conditional generation on ImageNet 256×256 using \method-$4$ with SiT-XL$/2$.
}
\label{fig:main:qual}
\vspace{-1.5em}
\end{figure*}

\smallskip 
\noindent
\textbf{Cross-condition generation.} To further validate the role of cross-sample attention, we conduct a controlled experiment by replacing one image in the group with a sample from a different class while keeping the latent variables fixed. We first generate a group of reference images and rank them by their cross-attention weights to the first sample, $p^{x_i \to x_1}$. Then, we gradually replace the condition of one sample with another class during the entire denoising process and show the results in Figure~\ref{fig:main:control_class}. 
We observed that the generation of the reference (green box) image is highly sensitive to which sample is replaced. When we replace a sample that originally receives high attention weights, the reference image changes significantly. 
In contrast, replacing a low-attention sample results in almost no visual difference. This indicates that cross-sample attention controls the inter-image correspondence within the group, with high-attention samples contributing more to the final generation, consistent with our earlier observations of cross-sample attention patterns.
Furthermore, we believe this property points to a promising future direction. When the group size is sufficiently large, the generation process of \method could be extended to handle diverse or cross-conditioned inputs, enabling more flexible inter-image correspondence within the generation process.

\begin{table}[t]
\centering
\resizebox{0.95\linewidth}{!}{
\begin{tabular}{lccccc}
\toprule
\multirow{2}{*}{Method}   & \multirow{2}{*}{Epoch} & \multicolumn{4}{c}{w/ CFG}        \\
\cmidrule(l){3-6}
                         &                                                   & FID  & IS    & Pre. & Rec. \\
\midrule
\multicolumn{4}{l}{\textcolor{gray}{\textit{\textbf{With semantic feature distillation}}}} \\


DDT-XL~\cite{wang2025ddt}                                        & 800                       & 1.26      & 310.6 & 0.79 & 0.65 \\
SiT-XL/2 + REPA-E~\cite{leng2025repae}                              &   800                     & 1.26  & 314.9 & 0.79 & \textbf{0.66} \\
SiT-XL/2 + REPA~\cite{repa}                                & 800                    & 1.42   & 305.7 & \textbf{0.80} & 0.64 \\
~ + our GroupDiff-4*                       & 800 + 100              & \textbf{1.14}          &  \textbf{315.3}     &  0.77    &  \textbf{0.66}    \\

\midrule
\multicolumn{4}{l}{\textcolor{gray}{\textit{\textbf{Without semantic feature distillation}}}} \\
ADM~\cite{dhariwal2021adm}                                           &  400                      & 4.59   & 186.7 & 0.82 & 0.52 \\
VDM++~\cite{kingma2023vdm}                                           & -                       & 2.12    & 267.7 & -    & -    \\
LDM-4~\cite{LDM}                                     &   1400                     & 3.60    & 247.7 & \textbf{0.87} & 0.48 \\
MDTv2-XL/2~\cite{gao2023mdtv2}                                  &   900                     & 1.58    & 314.7 & 0.79 & \textbf{0.65} \\
VAR-d30~\cite{tian2024var}                                      &  350                      & 1.92     & \textbf{323.1} & 0.82 & 0.59 \\
LlamaGen-3B~\cite{sun2024llamagen}                                  &  -                      & 2.18    & 263.3 & 0.81 & 0.58 \\
RandAR-XXL~\cite{pang2025randar}                                 &  300                      & 2.15  & \underline{322.0} & 0.79 & 0.62 \\
MaskDiT~\cite{zheng2023maskdit}                                       &   1600                     & 2.28     & 276.6 & 0.89 & 0.61 \\
\addlinespace
DiT-XL/2~\cite{peebles2023dit}                                   & 1400                   & 2.27    & 278.2 & \underline{0.83} & 0.57 \\
~ + \textbf{our} GroupDiff-$4$                          & 800        & 1.66        & 279.4      & \underline{0.83}    & 0.57     \\
~ + \textbf{our} GroupDiff-$4^{*}$                          & 1400 + 100              & \underline{1.55}      & 285.4      & 0.80     &  0.63  \\
SiT-XL/2~\cite{ma2024sit}                                      & 1400                   & 2.06  & 270.3 & 0.82 & 0.59 \\
~ + SRA~\cite{jiang2025sra}                                      & 800                    & 1.58    & 311.4 & 0.80 & 0.63 \\
~ + Dispersive Loss~\cite{wang2025disperseloss}                       & 1200                   & 1.97   & -     & -    & -    \\
~ + \textbf{our} GroupDiff-$4$                       & 800                    & 1.63       & 283.2      & 0.81      & \underline{0.64}     \\
~ + \textbf{our} GroupDiff-$4^{*}$                         & 1400 + 100             & \textbf{1.40}      & 290.7    & 0.79    & \underline{0.64}  \\
\bottomrule
\end{tabular}
}
\vspace{-0.5em}
\caption{\textbf{System-Level performance comparison} on ImageNet $256 \times 256$. Our \method enables the DiT/SiT model to achieve state-of-the-art performance both with/without semantic feature distillation.
$^{*}$: continue training from pre-trained checkpoint for an additional 100 epochs. 
}
\vspace{-1.5em}
\label{main:table:main_imagenet256}
\end{table}

\subsection{Benchmarking with Previous Systems}

We compare against leading generative systems in Table~\ref{main:table:main_imagenet256}. For this experiment, we train \method in two settings: from scratch and from the pre-trained weights, denoted as \method-$4$ and \method-$4^{*}$, respectively. When training from scratch, \method improves DiT-XL$/2$ with $29\%$ lower FID and SiT-XL$/2$ with $30\%$ lower FID while only using $57\%$ of original training iterations. 

For the second setting, we only use the $L_{group}$ as the training objective, no matter if other objectives, \eg $L_{repa}$ from REPA~\cite{repa}, exist in the previous stages. Notably, \method-$4$ with DiT-XL/2 achieves an FID of 1.55 (from 2.27) and \method-$4$ with SiT-XL/2 further improves to $1.40$ (from 2.06) with only 100 additional training epochs, outperforming all other state-of-the-art methods when no semantic feature distillation has been applied. Moreover, when using pre-trained weights from the semantic feature distillation method, \method again obtains a significant improvement, achieving an FID of 1.14 (down from 1.42). Qualitative samples are provided in Figure~\ref{fig:main:qual}.

\section{Discussion and Conclusion}
\noindent
\textbf{Limitations.}  While \method{} demonstrates strong improvements in generation quality, its increased training cost remains a challenge. 
When the group size is $n$, \method{}-\textit{f} and \method{}-\textit{l} require approximately $(n-1)\times$ and $(0.1n)\times$ longer training time in every iteration, and $(n-1)\times$ and $0.5(n-1)\times$ longer inference time, respectively.  Nevertheless, (a) this design opens a new avenue for exploring the trade-off between computational cost and generation quality, and (b) a high-quality model can serve as a teacher to distill faster and lighter students. We leave the study for a more efficient method for future exploration.

\smallskip
\noindent
\textbf{Conclusion.} We introduce Group Diffusion, a simple yet effective framework that reshapes diffusion training into a group-wise denoising process. By enabling cross-sample attention among related instances, the model implicitly learns relational structures that enhance representation quality and generation fidelity. Experiments on ImageNet demonstrate consistent FID improvements across architectures with minimal computational overhead. Beyond boosting performance, Group Diffusion provides a new lens connecting representation learning and generative modeling, suggesting that cross-sample interactions can serve as an implicit form of supervision for stronger and more generalizable diffusion models.

\clearpage

{
    \small
    \bibliographystyle{ieeenat_fullname}
    \bibliography{main}
}

\clearpage
\section*{Supplementary}
\begingroup
\hypersetup{linkcolor=black}

\startcontents
\printcontents{}{1}{}

\endgroup

\setcounter{section}{0}
\renewcommand\thesection{\Alph{section}}



\section{Implementation Details}

\subsection{Baselines}
We introduce the baselines of the leading generative systems as follows:
\begin{itemize}[leftmargin=1.5em]
    \item \textbf{ADM}~\cite{dhariwal2021adm} leverages classifier for guiding diffusion sampling to improve generation. 
    \item \textbf{LDM}~\cite{LDM} presents latent diffusion, enabling fast, high-resolution generation by training diffusion models in a latent space.
    \item \textbf{MDTv2}~\cite{gao2023mdtv2} combines masked token modeling with diffusion transformers to learn visual representations.
    \item \textbf{VAR}~\cite{tian2024var} introduces next-scale prediction to autoregressive generative models.
    \item \textbf{LlamaGen}~\cite{llamagen} shows vanilla autoregressive models could achieve strong generation performance at scale, outperforming diffusion baselines.
    \item \textbf{RandAR}~\cite{pang2025randar} proposes a decoder-only autoregressive model that utilizes position instruction tokens to generate image tokens in arbitrary orders.
    \item \textbf{MaskDiT}~\cite{zheng2023maskdit} uses masked input patches and an asymmetric encoder-decoder to achieve faster diffusion model training.
    \item \textbf{DiT}~\cite{peebles2023dit} proposes a scalable transformer architecture based on AdaIN-zero for diffusion model training. 
    \item \textbf{SiT}~\cite{ma2024sit} further improves the efficiency and scalability on DiT by introducing flow matching. 
    \item \textbf{REPA}~\cite{repa} analyzes the alignment between feature quality and generation fidelity of diffusion backbone and accelerates diffusion model training by aligning diffusion feature with pre-trained vision encoders. 
    \item \textbf{REPA-E}~\cite{leng2025repae} enables representation learning inside diffusion backbones by unlocking the latent encoder. 
    \item \textbf{DDT}~\cite{wang2025ddt} proposes a diffusion architecture that separates semantic encoding from high-frequency decoding to accelerate convergence during training.
    \item \textbf{SRA}~\cite{jiang2025sra} introduces a simple approach to align cross-layer diffusion backbone features to improve training efficiency without a pre-trained vision encoder. 
    \item \textbf{Dispersive Loss}~\cite{wang2025disperseloss} introduces a simple regularization loss that encourages internal representations to disperse in the hidden space to improve diffusion model training.

\end{itemize}

\begin{table*}[]
\centering

\resizebox{0.85\linewidth}{!}{

\begin{tabular}{lccccc}

\toprule
                   & \multicolumn{2}{c}{DiT-XL/2} & \multicolumn{2}{c}{SiT-XL/2} & SiT-XL/2-REPA \\
                   \cmidrule(l){2-3} \cmidrule(l){4-5} \cmidrule(l){6-6}
                   & GroupDiff-4   & GroupDiff-4* & GroupDiff-4   & GroupDiff-4* & GroupDiff-4*  \\
\midrule
\textbf{Architecture}       &               &              &               &              &               \\
Input dim.         & $32\times 32\times$ 4       & $32\times 32\times 4$      & $32\times 32\times 4$       & $32\times 32\times 4$      & $32\times 32\times 4$       \\
Num. layers        & 28            & 28           & 28            & 28           & 28            \\
Hidden dim.        & 1,152         & 1,152        & 1,152         & 1,152        & 1,152         \\
Num. heads         & 16            & 16           & 16            & 16           & 16            \\
\midrule
\textbf{Optimization}       &               &              &               &              &               \\
Resume             & -           & DiT-XL/2-7M        & -           & SiT-XL/2-7M          & REPA-4M           \\
Training Iteration & 4M            & 500K         & 4M            & 500K         & 500K          \\
Batch Size         & 256           & 256          & 256           & 256          & 256           \\
Optimzier          & AdamW         & AdamW        & AdamW         & AdamW        & AdamW         \\
lr                 & 0.0001        & 0.0001       & 0.0001        & 0.0001       & 0.0001        \\
betas              & (0.9, 0.999)  & (0.9, 0.999) & (0.9, 0.999)  & (0.9, 0.999) & (0.9, 0.999)  \\
weight decay       & 0.01          & 0.01         & 0.01          & 0.01         & 0.01          \\
\midrule
\textbf{GroupDiff}          &               &              &               &              &               \\
Mode               & GroupDiff-\textit{l}   & GroupDiff-\textit{l}  & GroupDiff-\textit{l}   & GroupDiff-\textit{l}  & GroupDiff-\textit{l}   \\
Query Method       & CLIP-L        & CLIP-L       & CLIP-L        & CLIP-L       & CLIP-L        \\
$\tau_{img}$         &  0.7             & 0.7              &   0.7            &  0.7            & 0.7              \\
Group Size         & 4             & 4            & 4             & 4            & 4             \\
Noise Var.         & 50            & 50           & 50            & 50           & 0             \\
\midrule
\textbf{Inference }         &               &              &               &              &               \\
Steps              & 250           & 250          & 250           & 250          & 250           \\
Guidance Scale     & 1.70          & 1.60      & 2.35           & 1.85          & 2.575           \\
Guidance Interval  & (0,1)         & (0,1)        & (0.25,1.0)         & (0.15,1.0)        & (0.25,0.75)        \\
\bottomrule

\end{tabular}

}
\caption{\textbf{Hyperparameter setup.}}
\label{tab_sup_hyper}
\end{table*}

\subsection{Evaluation Metric}
We use the conventional evaluation pipelines for class-conditional generative models, following ADM~\cite{dhariwal2021adm}. Specifically, we introduce the focusing concept of each metric: 

\begin{itemize}[leftmargin=1.5em]
    \item \textbf{Fr\'echet Inception Distance~(FID)}~\cite{heusel2017fid} evaluates the feature distance of generated images and the reference samples. Lower FID usually suggests better generation fidelity and diversity. 
    \item \textbf{Inception Score~(IS)}~\cite{salimans2016is_score} measures image quality and diversity based on how confidently a classifier recognizes each image and how varied the generated classes are. A higher Inception Score indicates a more meaningful image within each class. 
    \item  \textbf{Precision and recall}~\cite{kynkaanniemi2019improved_precision_and_recall}. Precision captures the realism of generated images, while recall captures their diversity relative to real data.
\end{itemize}


\subsection{Hyperparameter}

In Table~\ref{tab_sup_hyper},we introduce the hyperparameter setting for models reported at Table~3.

\vspace{-0.25em}
\section{Experiment}



\subsection{Ablations}

\begin{figure}[!h]
\centering 
\vspace{-0.5em}
\includegraphics[page=1,width=0.8\linewidth]{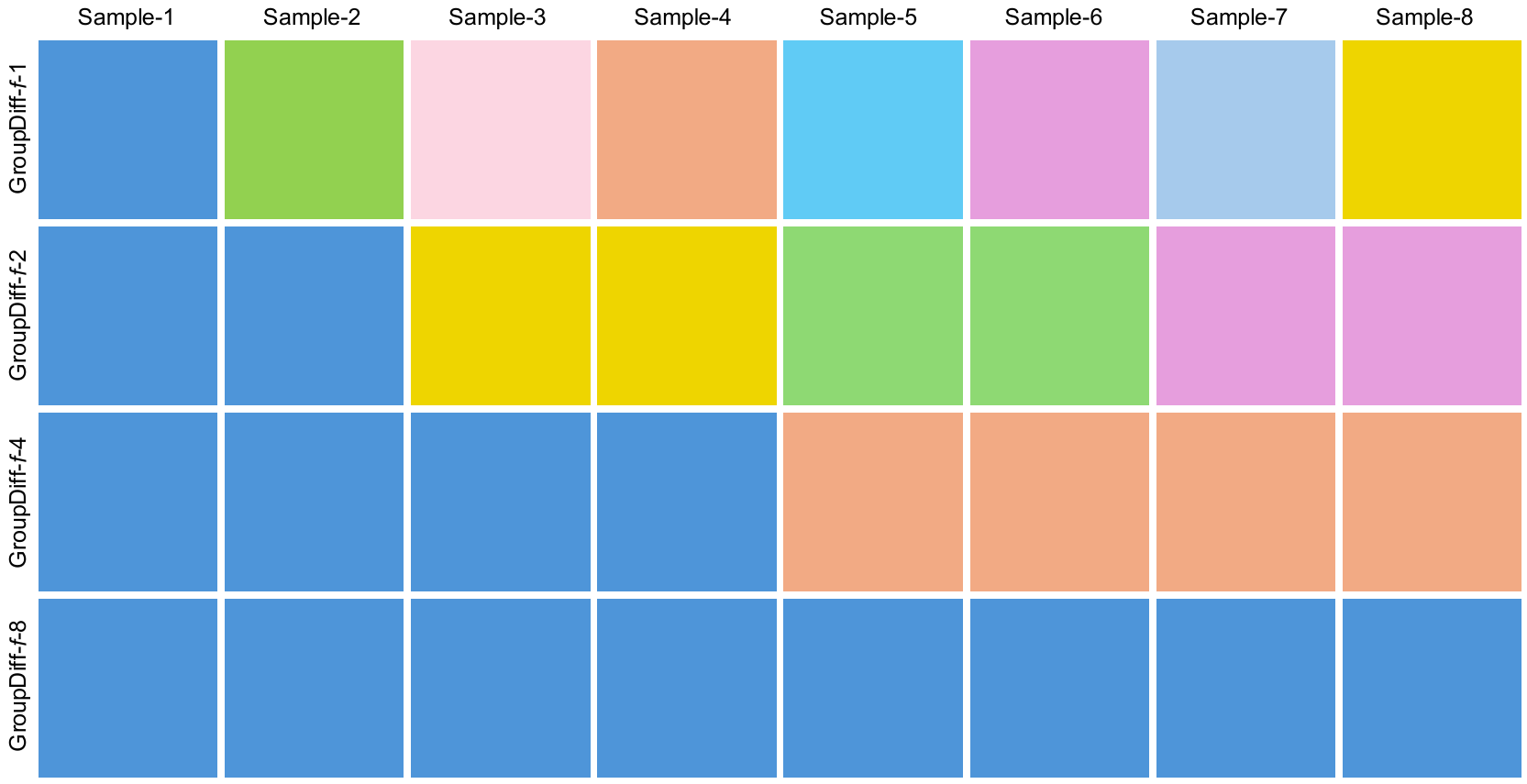}
\caption{\textbf{Group attention illustration.} 
In each row, samples in the sample group shares the same color block. 
 }
\label{fig:supp:attn}
\end{figure}

\begin{figure*}[!h]
\centering 
\vspace{-1em}
\includegraphics[page=1,width=1.0\linewidth]{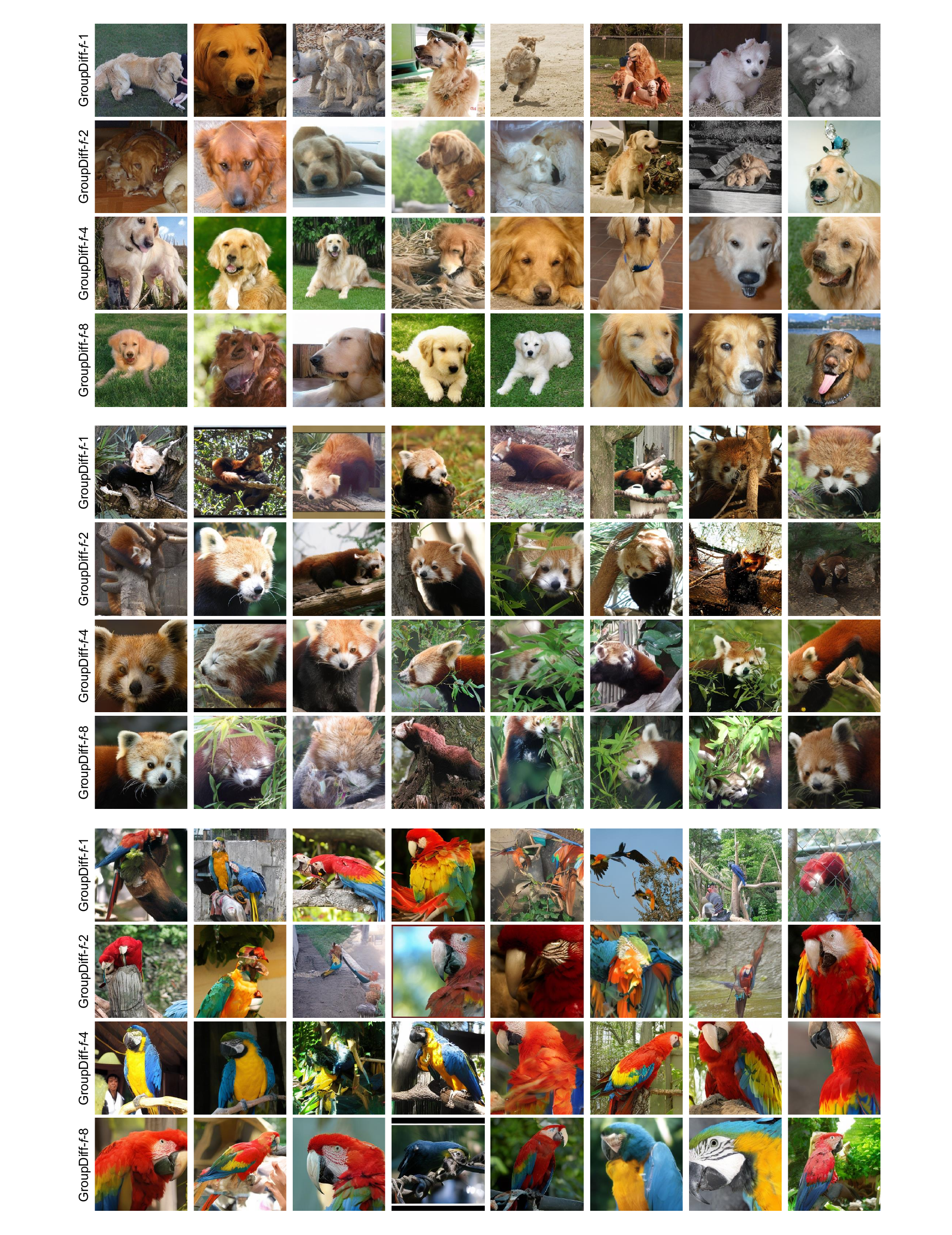}
\vspace{-3em}
\caption{\textbf{Uncurated generation results of \method-\textit{f} without classifier-free guidance.} Examples of class-conditional generation on ImageNet 256×256. \method with a larger group size consistently obtains better generation fidelity. }
\label{fig:supp:compare1}
\vspace{-1.5em}
\end{figure*}

\begin{figure*}[!h]
\centering 
\vspace{-0.5em}
\includegraphics[page=2,width=1.0\linewidth]{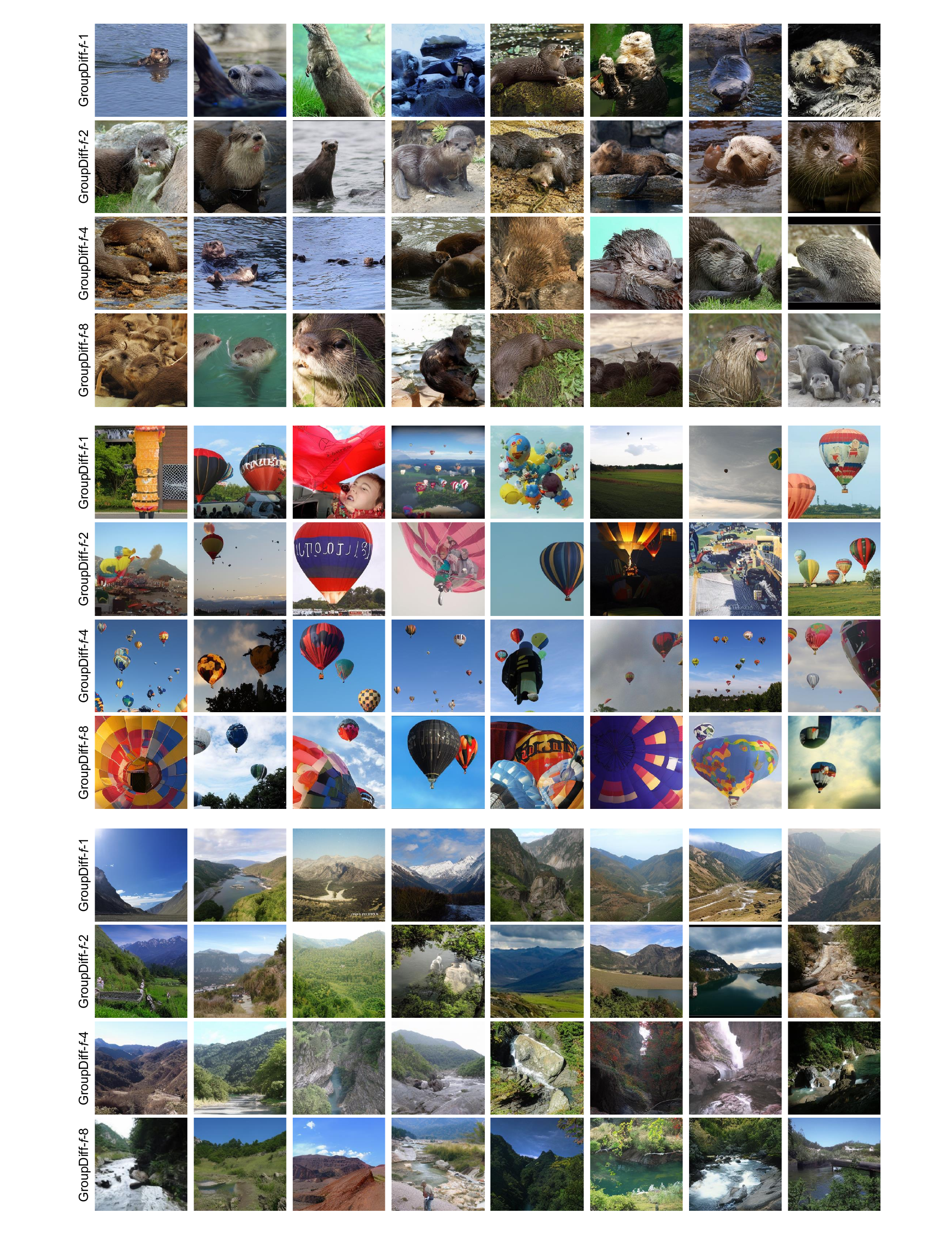}
\vspace{-3em}
\caption{\textbf{Uncurated generation results of \method-\textit{f} without classifier-free guidance.} Examples of class-conditional generation on ImageNet 256×256.\method with a larger group size consistently obtains better generation fidelity. }
\label{fig:supp:compare2}
\vspace{-1.5em}
\end{figure*}

\noindent
\textbf{\method-\textit{f}: group size.} We additionally investigate into the group size in \method-\textit{f} setting. Figure~\ref{fig:supp:attn} shows the which images shares the same group during inference. We compare the uncurated samples from \method-\textit{f}-\{1,2,3,4\} in Figure~\ref{fig:supp:compare1} and Figure~\ref{fig:supp:compare2}.
Our observation on \method-\textit{f} aligns that of \method-\textit{l}, where increasing the group size considerably improves the generation fidelity.

\smallskip 
\noindent
\textbf{\method-\textit{l*} : query method}. Beyond training from scratch, resuming from individual diffusion offers an efficient solution to adding \method over existing pipelines. Thus, we also explore different query methods under this setting. Table~\ref{tab:supp:abl_query} shows CLIP-L yields the optimality performance while the simplest \method-4$^*$ obtains a considerable improvement (14.5\%) over the baseline, highlighting the effectiveness of cross-sample attention.

\begin{table}[]
\resizebox{\linewidth}{!}{
\begin{tabular}{lccccc}
\toprule
Method       & Query Method & FID~$\downarrow$  & IS~$\uparrow$    & Pre.$\uparrow$ & Rec.$\uparrow$ \\
\midrule
SiT-XL/2     & -            & 2.06 & 270.3 & 0.82 & 0.59 \\
~ + GroupDiff-4* & Class        & 1.76 & 283.5 & 0.81 & 0.61 \\
~ + GroupDiff-4* & CLIP-L       & 1.40 & 290.7 & 0.79 & 0.64 \\
\bottomrule
\end{tabular}
}
\caption{\textbf{Ablation: query method}.
$^{*}$: continue training from pre-trained checkpoint for an additional 100 epochs. 
} 
\label{tab:supp:abl_query}
\end{table}

\subsection{Extending to Pixel Diffusion.} We further validate GroupDiff on pixel diffusion systems. As shown in Table~\ref{tab:supp:pixel}, \method-4 with JiT-B/16 delivers a substantial $15.8\%$ improvement with only 100 additional training steps when resumed from a pre-trained model. This again highlights the effectiveness of cross-sample collaboration in pixel diffusion and its strong potential for broader applicability.

\begin{table}[]
\centering
\begin{tabular}{llll}
\toprule
Method                  & params & FID  & IS    \\
\midrule
ADM-G~\cite{dhariwal2021adm}                   & 559M   & 7.72 & 172.7 \\
RIN~\cite{jabri2022rin}& 320M   & 3.95 & 216   \\
SiD~\cite{zhou2024sid}, UViT/2             & 2B     & 2.44 & 256.3 \\
PixelFlow~\cite{chen2025pixelflow}, XL/4          & 677M   & 1.98 & 282.1 \\
PixNerd~\cite{wang2025pixnerd},  XL/16         & 700M   & 2.15 & 297   \\
JiT-H/16~\cite{li2025jit}                & 953M   & 1.86 & 303.4 \\
JiT-B/16~\cite{li2025jit}                & 131M   & 3.66 & 275.1 \\
~ + \textbf{our} GroupDiff-4* & 131M   & 3.08 & 245.6 \\
\bottomrule
\end{tabular}
\caption{\textbf{System-level performance of pixel diffusion models} evaluated on ImageNet 256$\times$256. $^{*}$: continue training from pre-trained checkpoint for an additional 100 epochs.}
\label{tab:supp:pixel}
\end{table}
\subsection{Additional Qualitative Results. }
\begin{figure*}[!h]
\centering 
\includegraphics[page=1,width=0.98\linewidth]{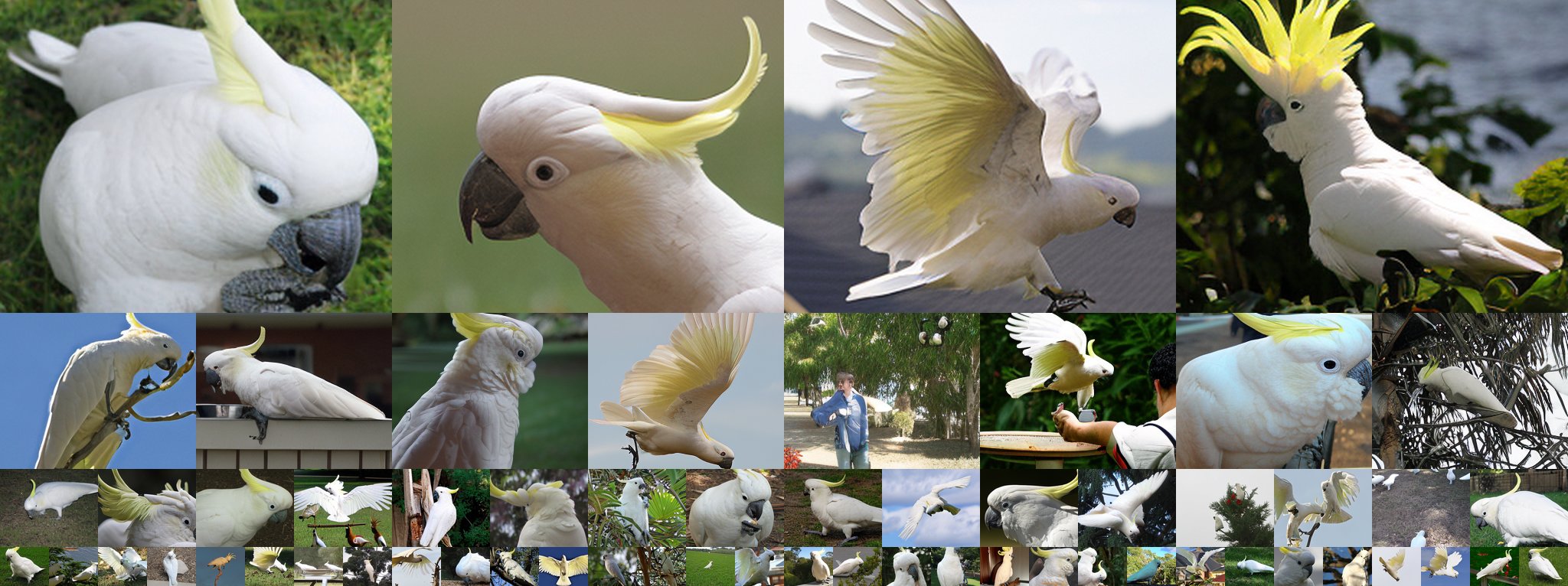}
\vspace{-0.5em}
\caption{\textbf{Uncurated generation results of \method-\textit{4}}. We use classifier-free guidance
with w= 3.5. Class label = “sulphur-crested cockatoo, Kakatoe galerita, Cacatua galerita” (89). }

\label{fig:supp:sample_start}
\end{figure*}

\begin{figure*}[!h]
\centering 
\includegraphics[page=1,width=0.98\linewidth]{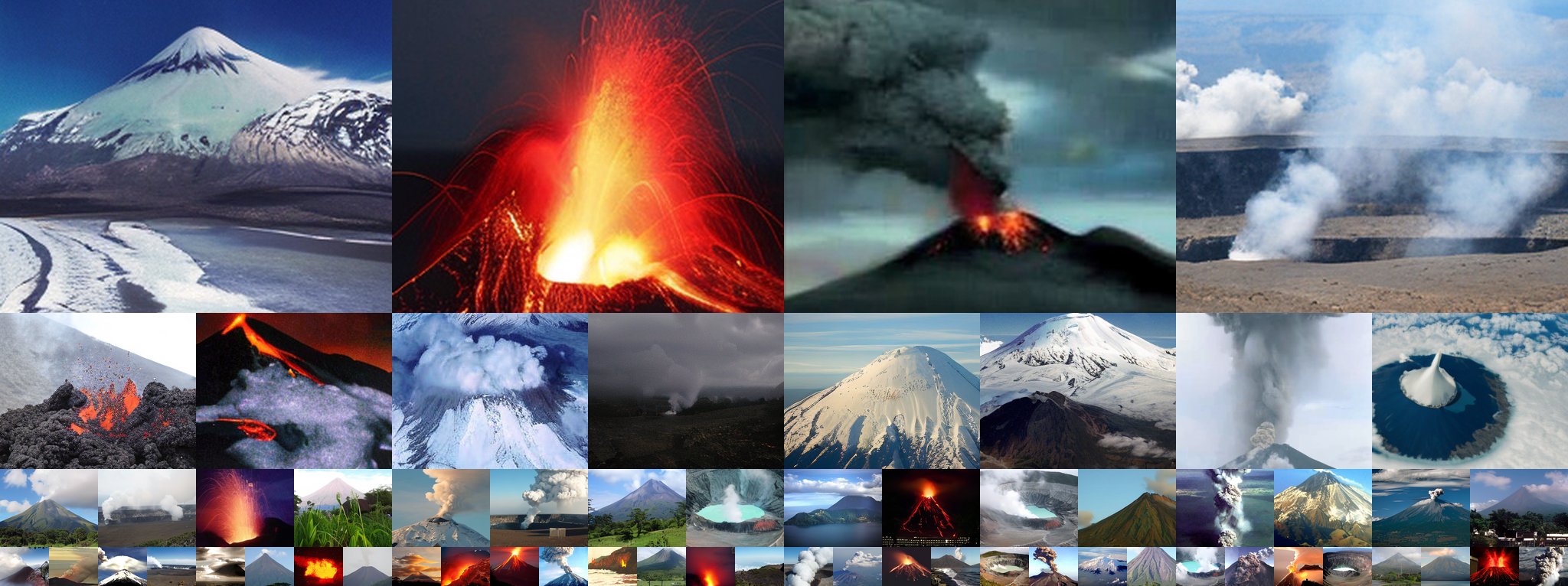}
\vspace{-0.5em}
\caption{\textbf{Uncurated generation results of \method-\textit{4}}. We use classifier-free guidance
with w= 3.5. Class label = “volcano” (980). }
\label{fig:supp:sample_end}
\end{figure*}

We provide additional uncurated samples generated by \method-4 in Figures~\ref{fig:supp:sample_start}–\ref{fig:supp:sample_end}.

\clearpage

\begin{figure*}[!h]
\vspace{-1em}
\centering 
\includegraphics[page=1,width=0.98\linewidth]{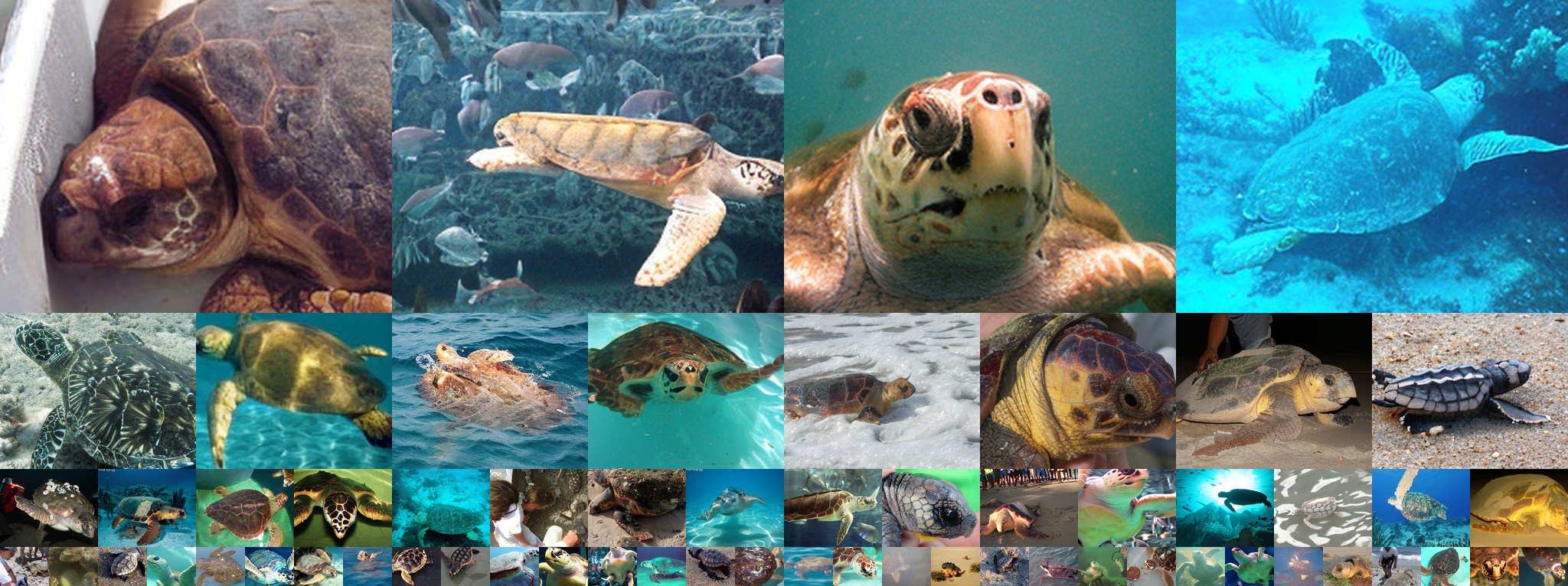}
\vspace{-0.5em}
\caption{\textbf{Uncurated generation results of \method-\textit{4}}. We use classifier-free guidance
with w= 3.5. Class label = “loggerhead sea turtle” (33). }
\end{figure*}

\begin{figure*}[!h]
\centering 
\includegraphics[page=1,width=0.98\linewidth]{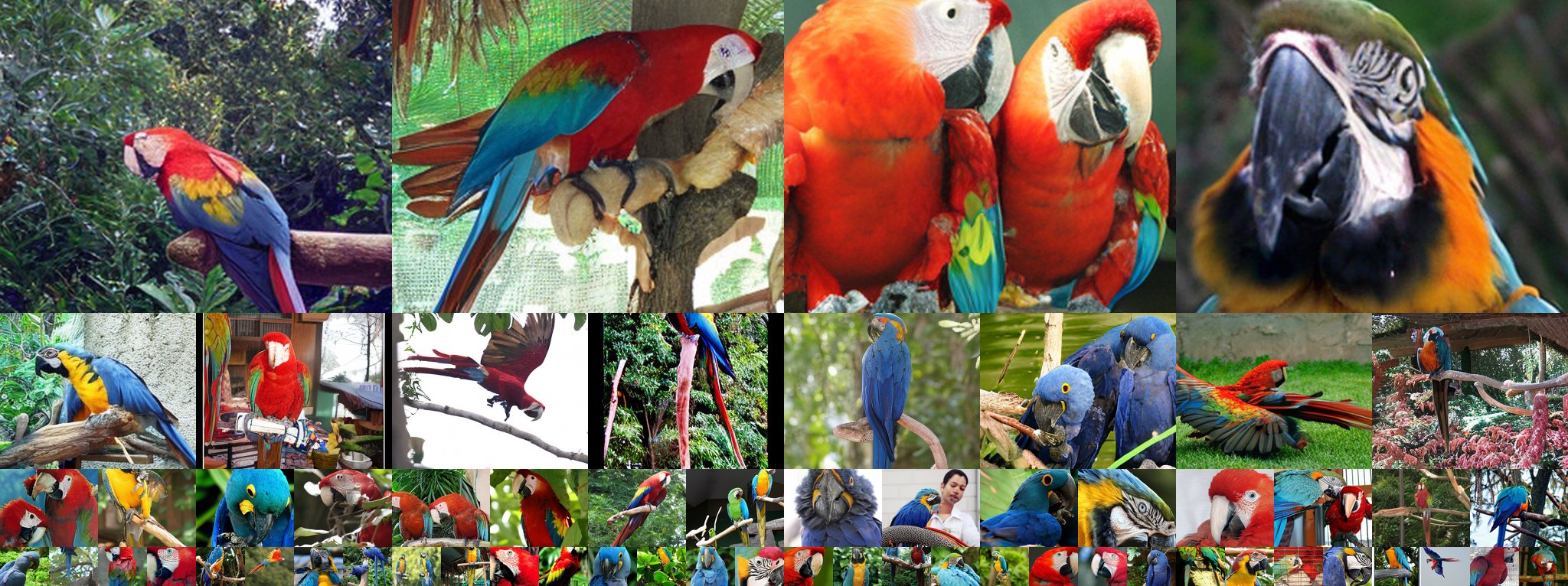}
\vspace{-0.5em}
\caption{\textbf{Uncurated generation results of \method-\textit{4}}. We use classifier-free guidance
with w= 3.5. Class label = “macaw” (88). }
\end{figure*}

\vspace{-2.5em}

\clearpage

\begin{figure*}[!h]
\centering 
\vspace{-1em}
\includegraphics[page=1,width=0.98\linewidth]{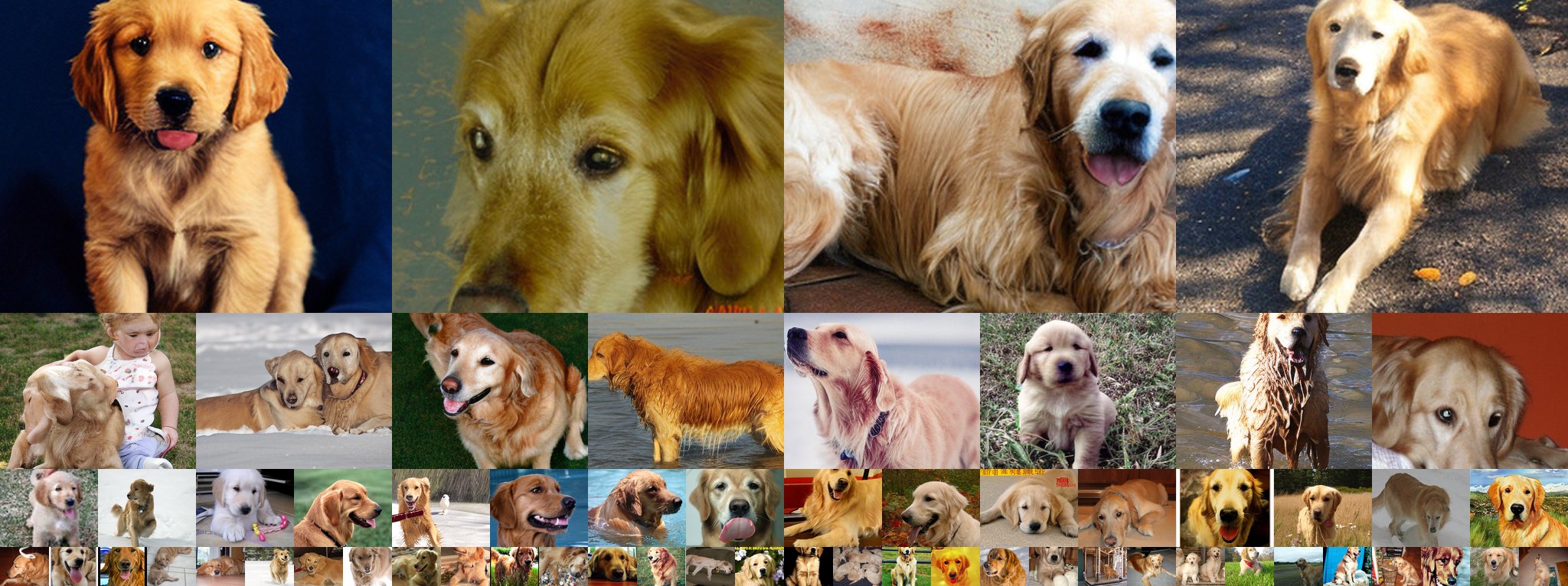}
\vspace{-0.5em}
\caption{\textbf{Uncurated generation results of \method-\textit{4}}. We use classifier-free guidance
with w= 3.5. Class label = “golden retriever” (207). }
\end{figure*}

\begin{figure*}[!h]
\centering 
\includegraphics[page=1,width=0.98\linewidth]{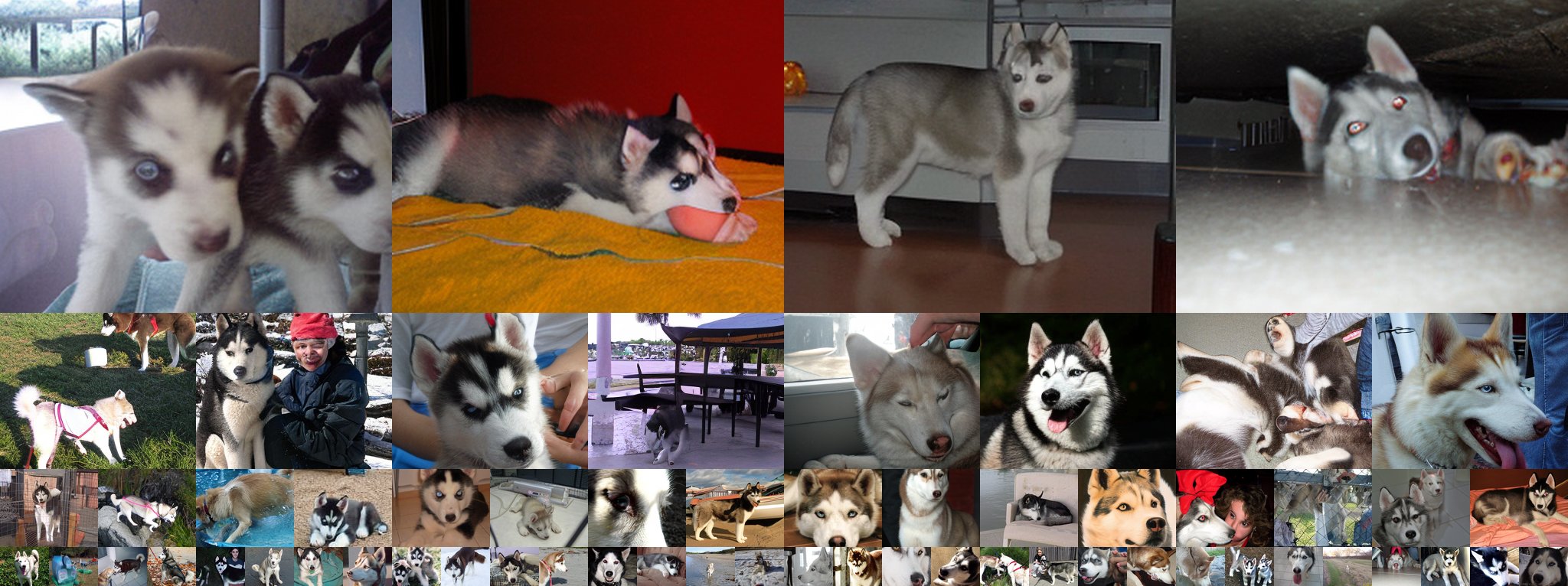}
\vspace{-0.5em}
\caption{\textbf{Uncurated generation results of \method-\textit{4}}. We use classifier-free guidance
with w= 3.5. Class label = “Siberian husky” (250). }
\end{figure*}

\begin{figure*}[!h]
\centering 
\includegraphics[page=1,width=0.98\linewidth]{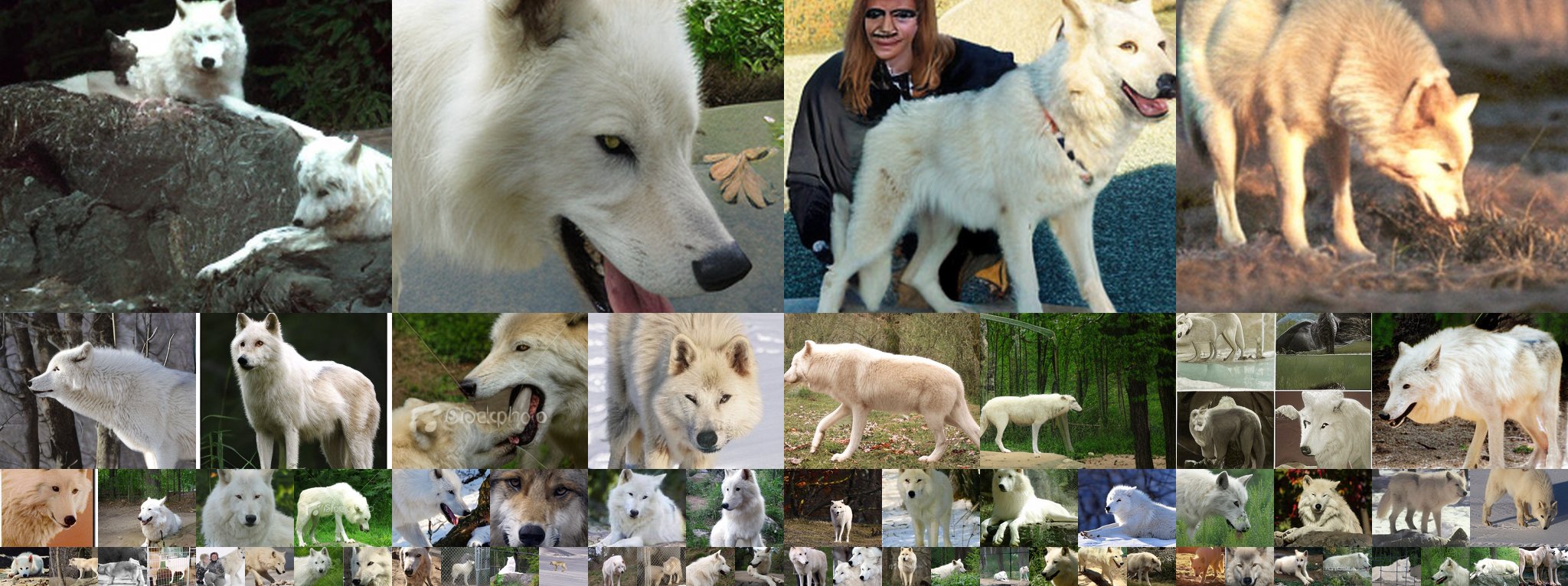}
\vspace{-0.5em}
\caption{\textbf{Uncurated generation results of \method-\textit{4}}. We use classifier-free guidance
with w= 3.5. Class label = “white wolf, Arctic wolf, Canis lupus tundrarum” (270). }
\end{figure*}

\clearpage

\begin{figure*}[!h]
\centering 
\vspace{-1em}
\includegraphics[page=1,width=0.98\linewidth]{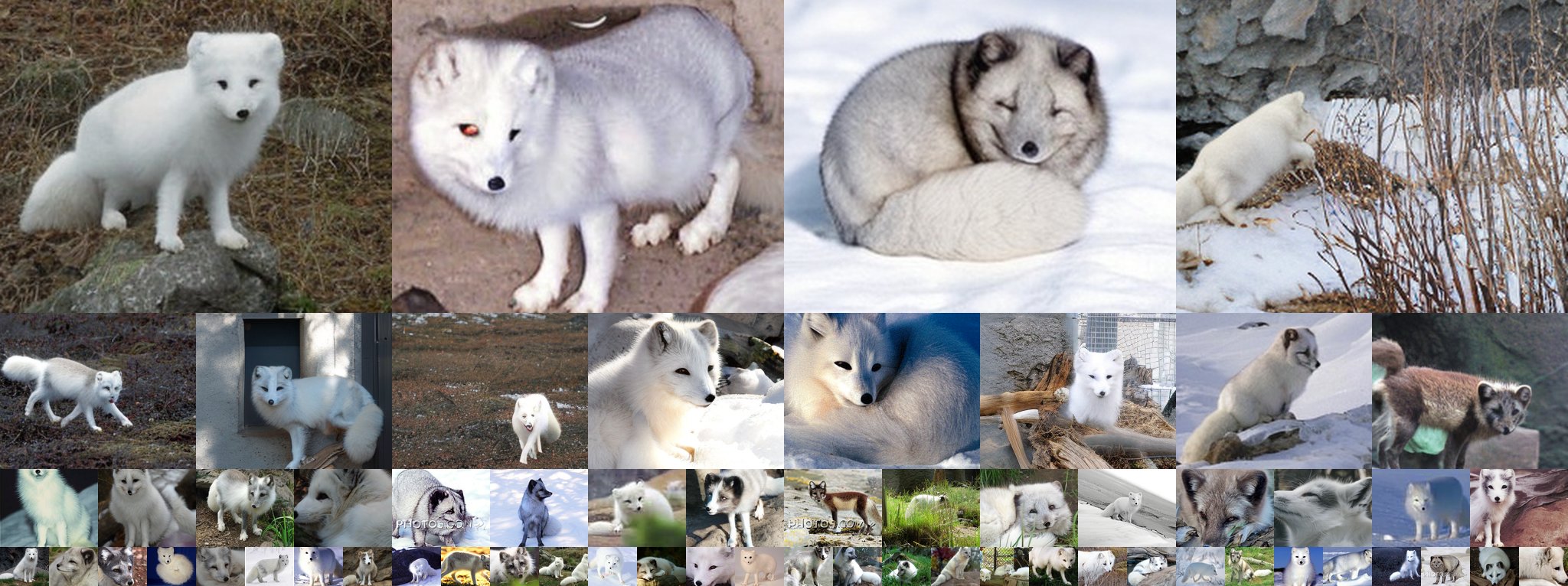}
\vspace{-0.5em}
\caption{\textbf{Uncurated generation results of \method-\textit{4}}. We use classifier-free guidance
with w= 3.5. Class label = “Arctic fox, white fox, Alopex lagopus” (279). }
\end{figure*}

\begin{figure*}[!h]
\centering 
\includegraphics[page=1,width=0.98\linewidth]{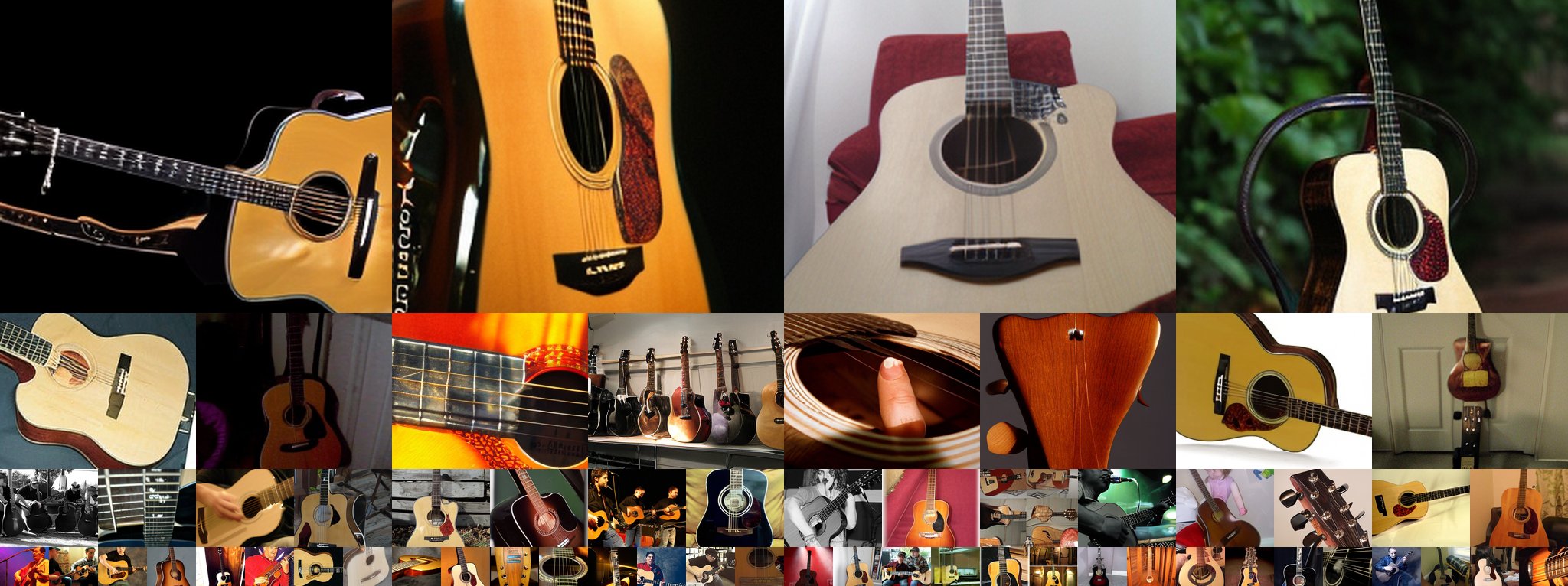}
\vspace{-0.5em}
\caption{\textbf{Uncurated generation results of \method-\textit{4}}. We use classifier-free guidance
with w= 3.5. Class label = “acoustic guitar” (402). }
\end{figure*}

\begin{figure*}[!h]
\centering 
\includegraphics[page=1,width=0.98\linewidth]{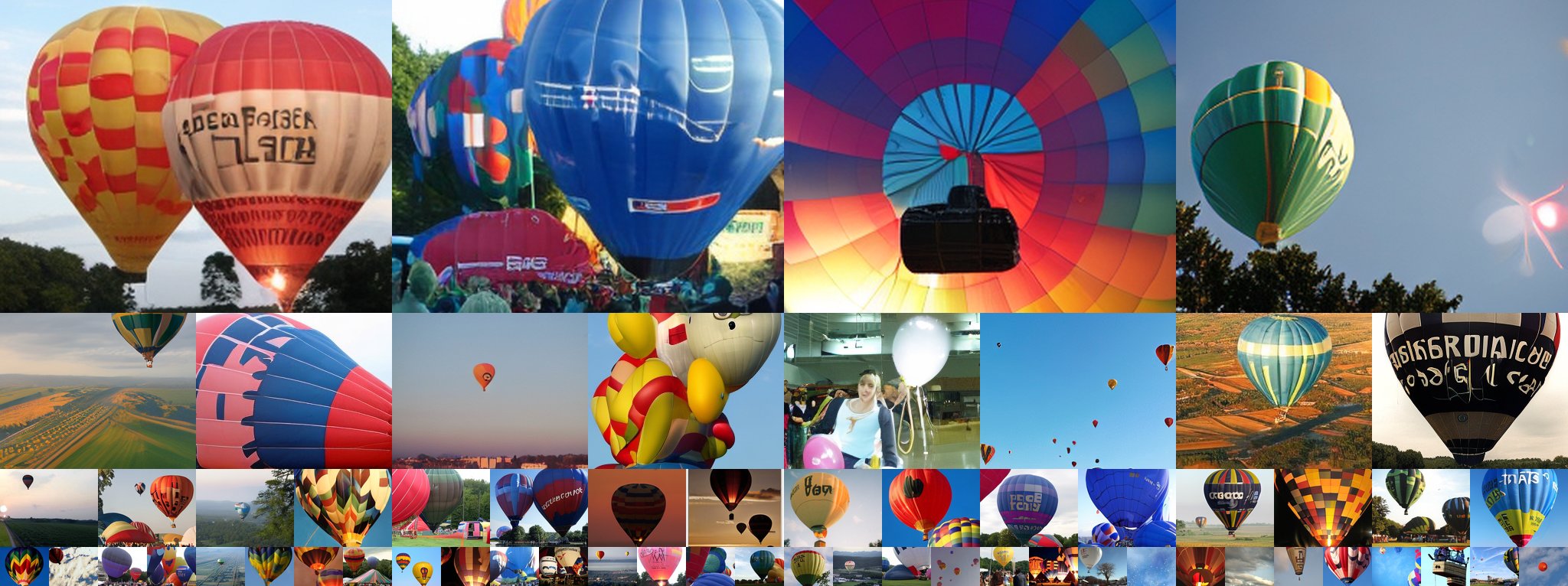}
\vspace{-0.5em}
\caption{\textbf{Uncurated generation results of \method-\textit{4}}. We use classifier-free guidance
with w= 3.5. Class label = “balloon” (417). }
\end{figure*}

\clearpage

\begin{figure*}[!h]
\centering 
\vspace{-1em}
\includegraphics[page=1,width=0.98\linewidth]{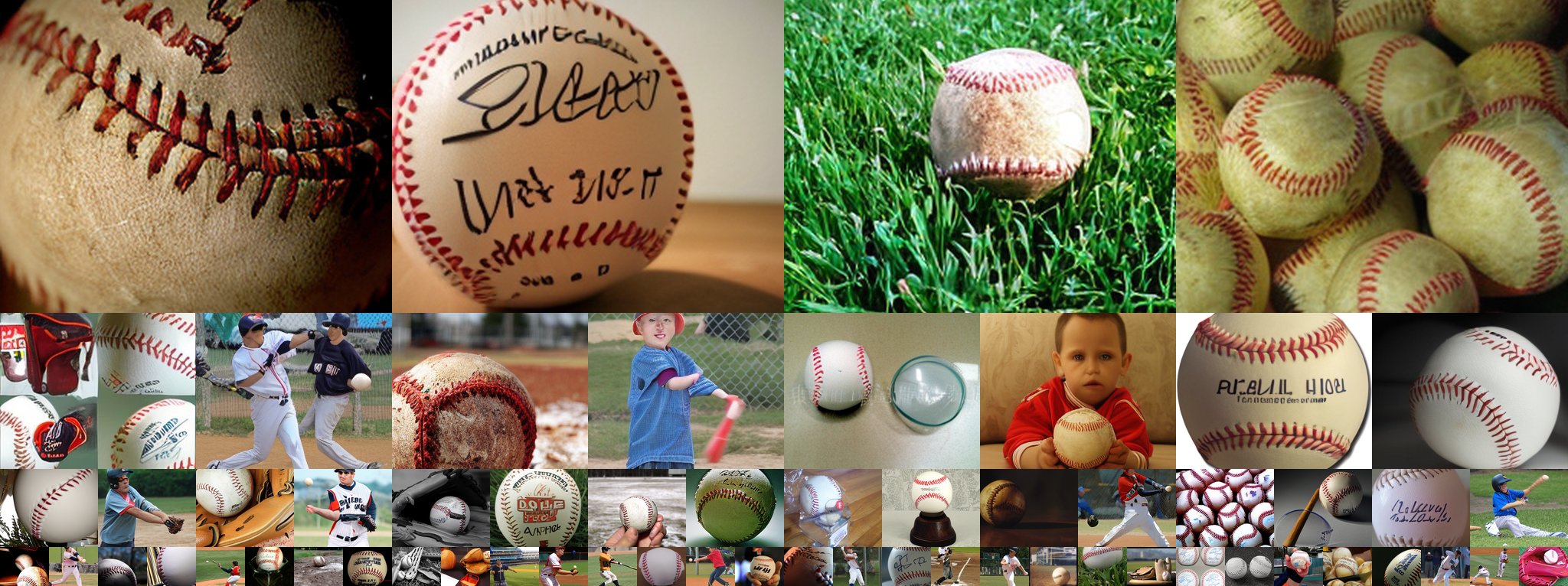}
\vspace{-0.5em}
\caption{\textbf{Uncurated generation results of \method-\textit{4}}. We use classifier-free guidance
with w= 3.5. Class label = “baseball” (429). }
\end{figure*}

\begin{figure*}[!h]
\centering 
\includegraphics[page=1,width=0.98\linewidth]{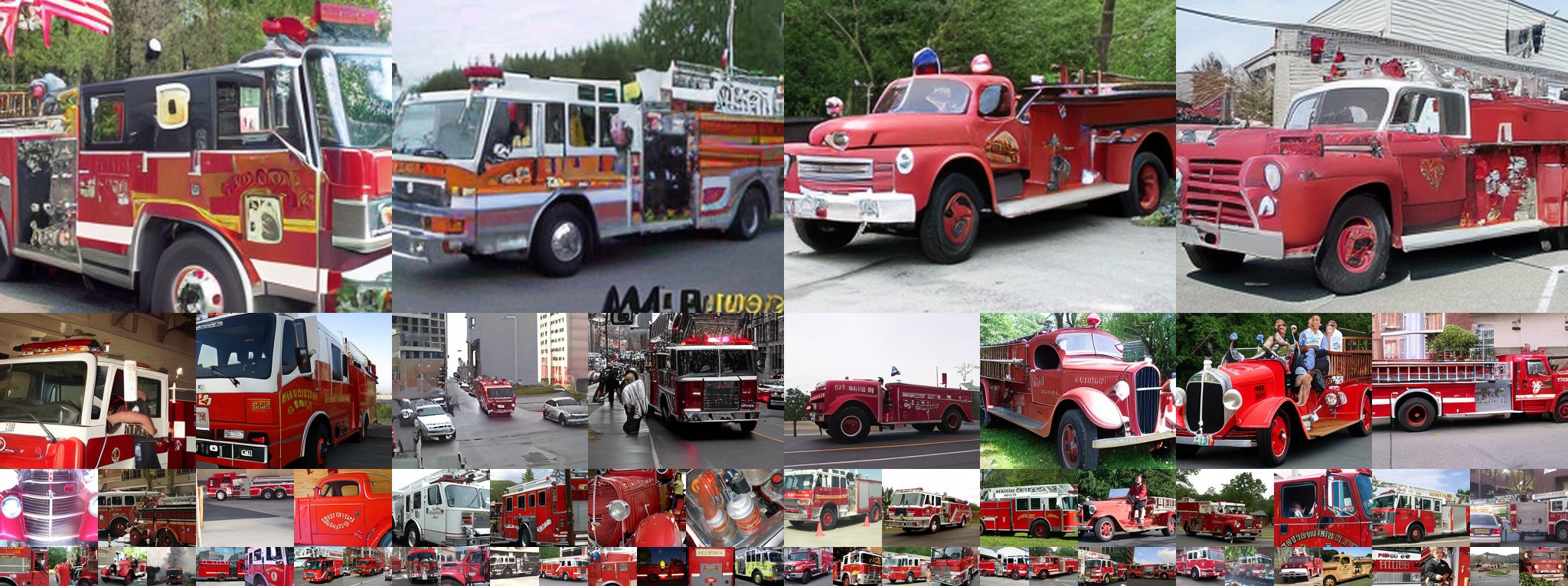}
\vspace{-0.5em}
\caption{\textbf{Uncurated generation results of \method-\textit{4}}. We use classifier-free guidance
with w= 3.5. Class label = “fire engine, fire truck” (555). }
\end{figure*}

\begin{figure*}[!h]
\centering 
\includegraphics[page=1,width=0.98\linewidth]{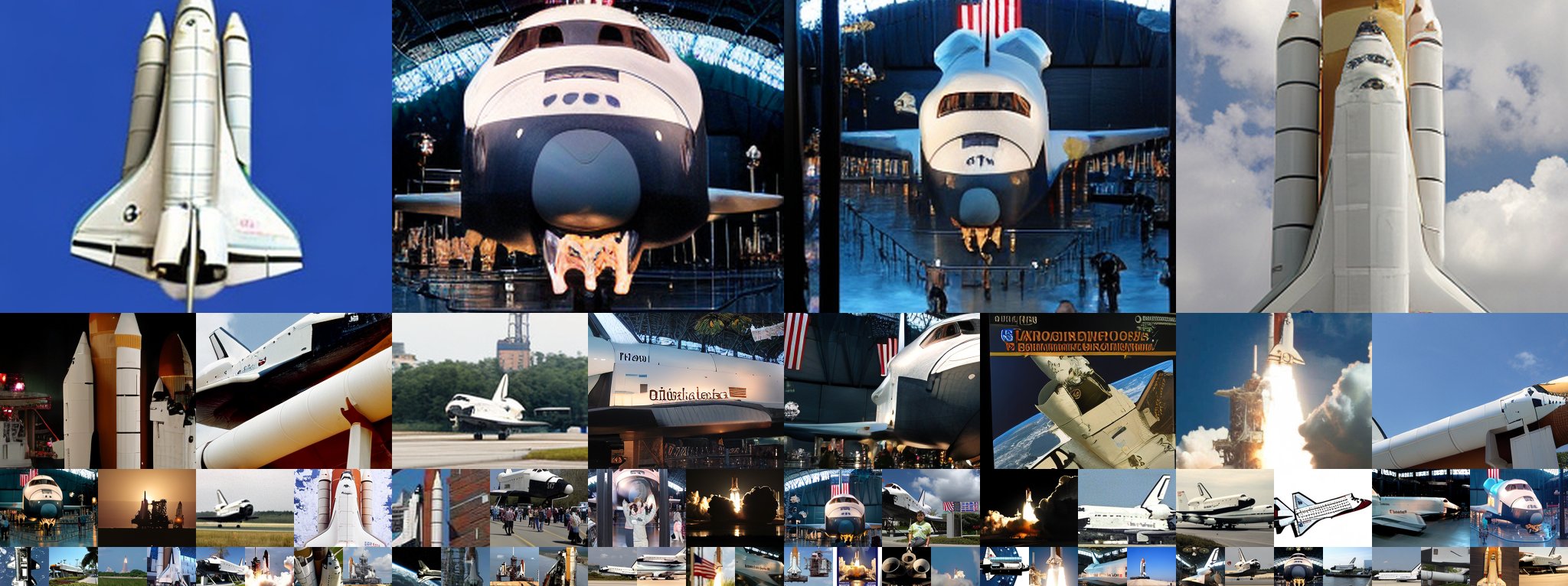}
\vspace{-0.5em}
\caption{\textbf{Uncurated generation results of \method-\textit{4}}. We use classifier-free guidance
with w= 3.5. Class label = “space shuttle” (812). }
\end{figure*}

\clearpage

\begin{figure*}[!h]
\centering 
\vspace{-1em}
\includegraphics[page=1,width=0.98\linewidth]{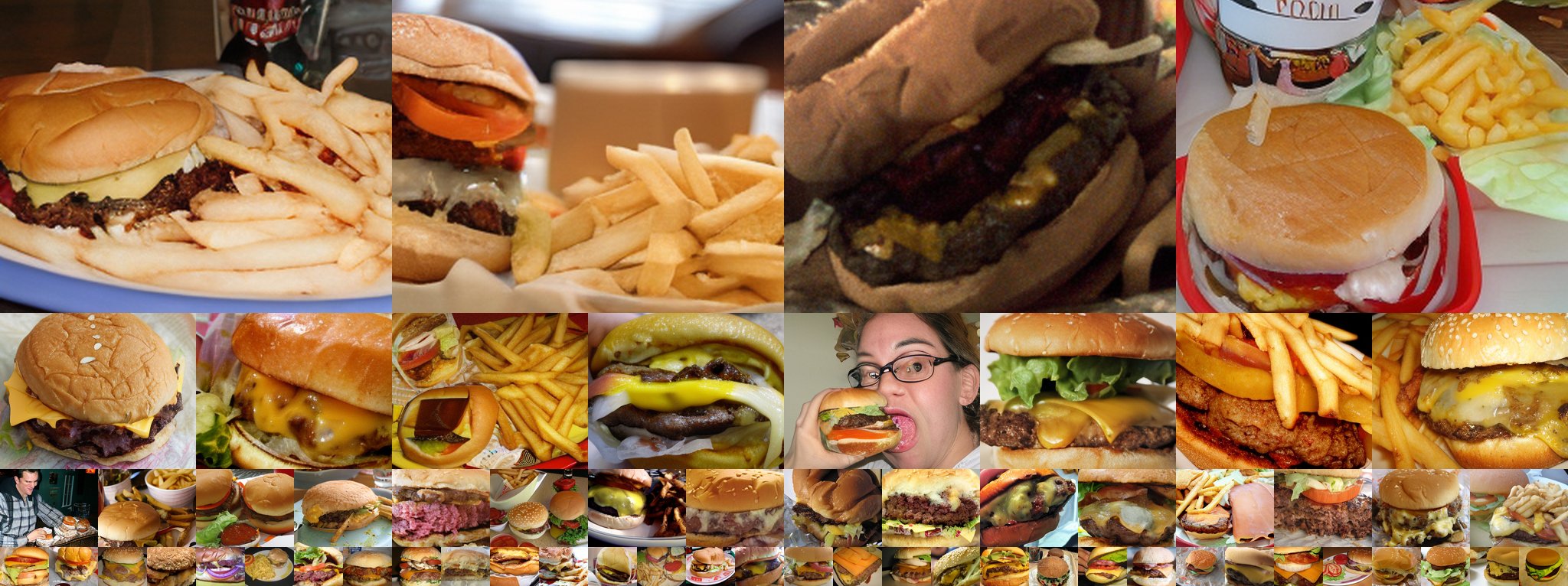}
\vspace{-0.5em}
\caption{\textbf{Uncurated generation results of \method-\textit{4}}. We use classifier-free guidance
with w= 3.5. Class label = “cheeseburger” (933). }
\end{figure*}

\begin{figure*}[!h]
\centering 
\includegraphics[page=1,width=0.98\linewidth]{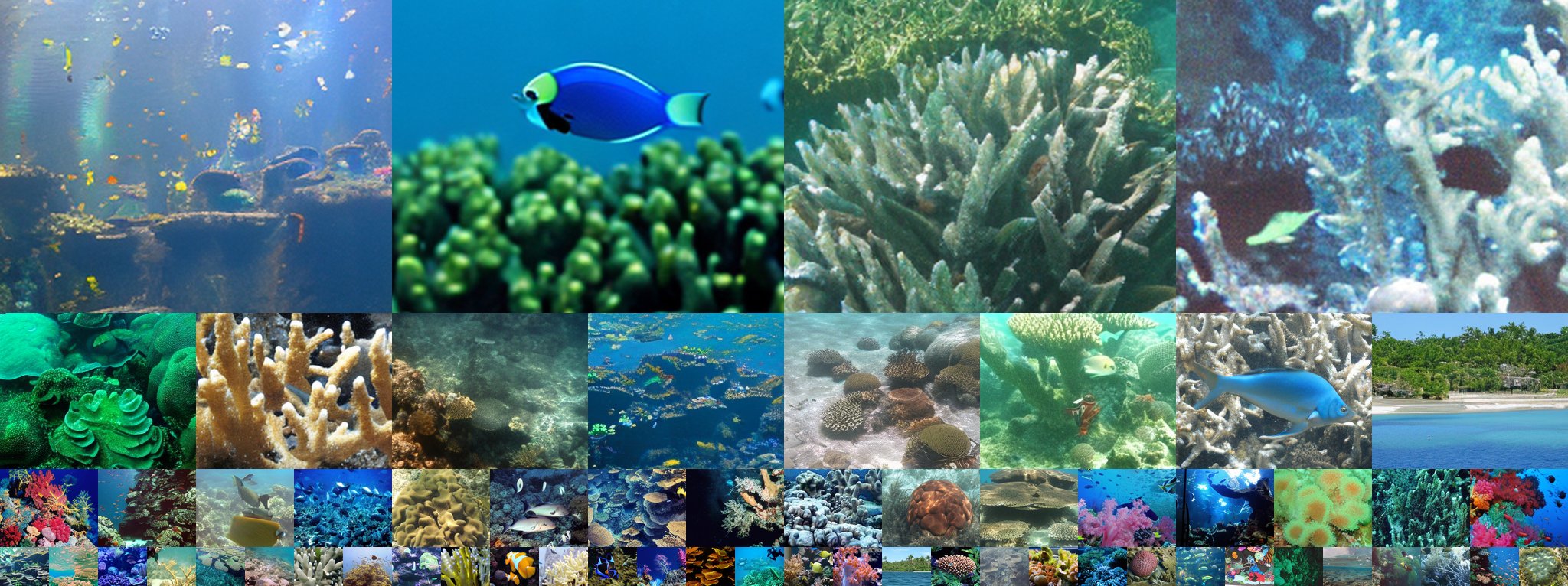}
\vspace{-0.5em}
\caption{\textbf{Uncurated generation results of \method-\textit{4}}. We use classifier-free guidance
with w= 3.5. Class label = “coral reef” (973). }
\end{figure*}

\clearpage

\subsection{Cross-Sample Score Visualization}
\begin{figure}[!h]
\centering 
\includegraphics[page=1,width=0.8\linewidth]{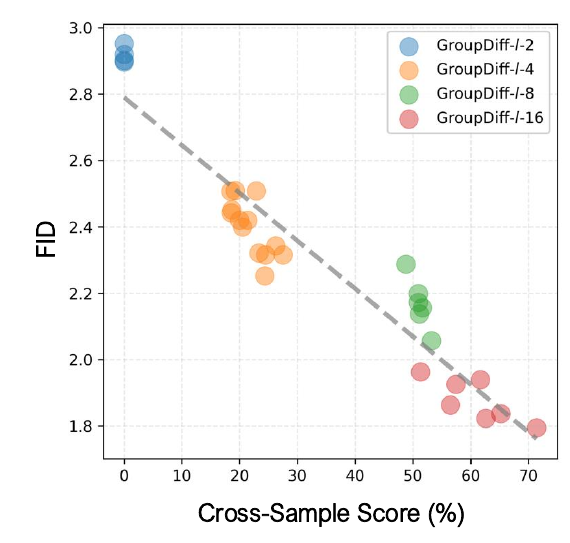}
\vspace{-0.5em}
\caption{
\textbf{FID vs Cross-Sample Score (group-level)} Our \method shows a strong correlation (0.94) between cross-attention to other samples and generation quality. }
\label{fig:supp:cross-sample-group}
\end{figure}

Additionally, we show the relation between FID and cross-sample score computed by the group-level mean and max of the attention score in Figure~\ref{fig:supp:cross-sample-group}. 

\subsection{Text-to-Image Generation}
\begin{table}[]
\centering
\begin{tabular}{lcc}
\toprule
Method              & Type      & FID   \\
\midrule
AttnGAN~\cite{xu2018attngan}             & GAN       & 35.49 \\
DM-GAN~\cite{zhu2019dm_gan}              & GAN       & 32.64 \\
VQ-Diffusion~\cite{gu2022vq_diffusion}        & Diffusion & 19.75 \\
DF-GAN~\cite{tao2022df_gan}              & GAN       & 19.32 \\
XMC-GAN~\cite{zhang2021xmc_gan}             & GAN       & 9.33  \\
Frido~\cite{fan2023frido}               & Diffusion & 8.97  \\
LAFITE~\cite{zhou2022lafite}              & GAN       & 8.12  \\
U-Net~\cite{bao2023uvit}               & Diffusion & 7.32  \\
U-ViT-S/2~\cite{bao2023uvit}           & Diffusion & 5.95  \\
U-ViT/S/2 (Deep)~\cite{bao2023uvit}    & Diffusion & 5.45  \\
MMDiT~\cite{stablediffusion_3}               & Diffusion & 5.3   \\
DiT-XL/2 w/ Cross-Attention~\cite{peebles2023dit} & Diffusion & 6.95 \\
~ + \textbf{our} \method-4 & Diffusion & 6.65 \\
\bottomrule
\end{tabular}
\vspace{-0.5em}
\caption{\textbf{Quantitative comparison} on text-to-image generation (MS-COCO).  }
\label{tab:main:t2i}
\end{table}
We also validate \method in text-to-image generation. We mostly follow the experimental setup used in
U-ViT~\cite{bao2023uvit} unless otherwise specified: we train the model from scratch on a train split
of the MS-COCO dataset and use a validation split for evaluation. We use DiT-XL/2 with Cross-Attention and train it for 150K iterations with a batch size
of 256. We use the frozen CLIP text encoder to extract text prompts from captions. Table~\ref{tab:main:t2i} shows that \method remains effective in the T2I generation setting without bells and whistles, highlighting the importance of applying cross-sample attention even with text conditions.



\end{document}